\pdfoutput=1
\documentclass[11pt]{article}

\usepackage[preprint]{acl}

\usepackage{times}
\usepackage{latexsym}

\usepackage[T1]{fontenc}
\usepackage[utf8]{inputenc}

\usepackage{microtype}

\usepackage{inconsolata}

\usepackage{graphicx}
\usepackage{array}
\usepackage{tabularx}
\usepackage{float}
\usepackage{multicol, multirow}
\usepackage{booktabs}
\usepackage{amsmath}
\usepackage{tcolorbox}
\usepackage{enumitem}
\usepackage{bookmark}
\usepackage{adjustbox}
\usepackage{makecell}
\usepackage{xurl}
\usepackage{colortbl}
\usepackage{tikz}
\usepackage{enumitem}
\usepackage{soul}

\usetikzlibrary{shapes.multipart, positioning}
\newcommand{\rom}[1]{\lowercase\expandafter{\romannumeral #1\relax}}

\newcommand{\name}{\textsc{\textsf{CORDIAL}}}

\definecolor{myblue}{RGB}{173, 216, 230} % Light blue background
\definecolor{myborder}{RGB}{0, 102, 204} % Dark blue border

\newcolumntype{Y}{>{\centering\arraybackslash}X}
\newtcbox{\myovalbox}{colback=cyan,boxrule=0pt,arc=2pt,
  boxsep=0pt,left=1pt,right=1pt,top=0pt,bottom=0pt}

\newcommand{\psulogo}{\raisebox{3.4pt}{\includegraphics[scale=0.025]{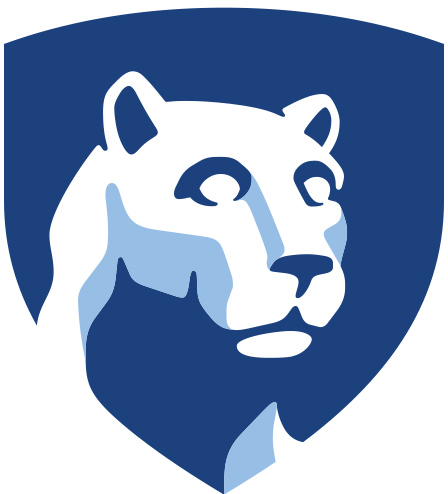}}}
\newcommand{\nittlogo}{\raisebox{3.4pt}{\includegraphics[scale=0.025]{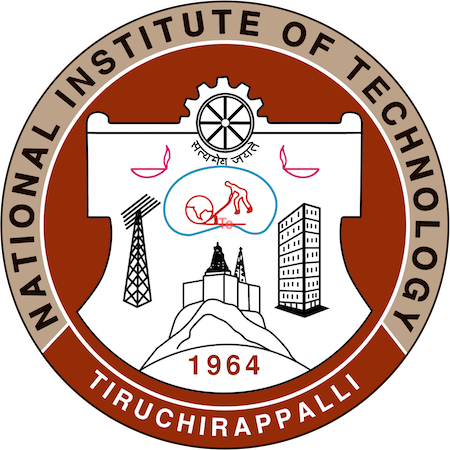}}}

\title{{\name}: Can Multimodal Large Language Models Effectively Understand Coherence Relationships?}

\author{
  \textbf{Aashish Anantha Ramakrishnan\psulogo},
  \textbf{Aadarsh Anantha Ramakrishnan\nittlogo},
  \textbf{Dongwon Lee\psulogo}
\\
  The Pennsylvania State University\textsuperscript{\psulogo},
  National Institute of Technology, Tiruchirappalli\textsuperscript{\nittlogo}
\\
  \texttt{
    \{aza6352, dul13\}@psu.edu\psulogo, 106121001@nitt.edu\nittlogo
  }
}

\begin{document}
\maketitle

\begin{abstract}
Multimodal Large Language Models (MLLMs) are renowned for their superior instruction-following and reasoning capabilities across diverse problem domains. However, existing benchmarks primarily focus on assessing factual and logical correctness in downstream tasks, with limited emphasis on evaluating MLLMs' ability to interpret pragmatic cues and intermodal relationships. To address this gap, we assess the competency of MLLMs in performing {\em Multimodal Discourse Analysis} (MDA) using Coherence Relations. Our benchmark, {\name}, encompasses a broad spectrum of Coherence Relations across 3 different discourse domains at varying levels of granularity. Through our experiments on 10+ MLLMs employing different prompting strategies, we show that even top models like Gemini 1.5 Pro and GPT-4o fail to match the performance of simple classifier-based baselines. This study emphasizes the need to move beyond similarity-based metrics and adopt a discourse-driven framework for evaluating MLLMs, providing a more nuanced assessment of their capabilities. The benchmark and code are available at: \url{https://aashish2000.github.io/CORDIAL/}.

\end{abstract}

\section{Introduction}

\begin{figure*}[!ht]
    \centering
    \frame{\includegraphics[width=0.85\linewidth]{"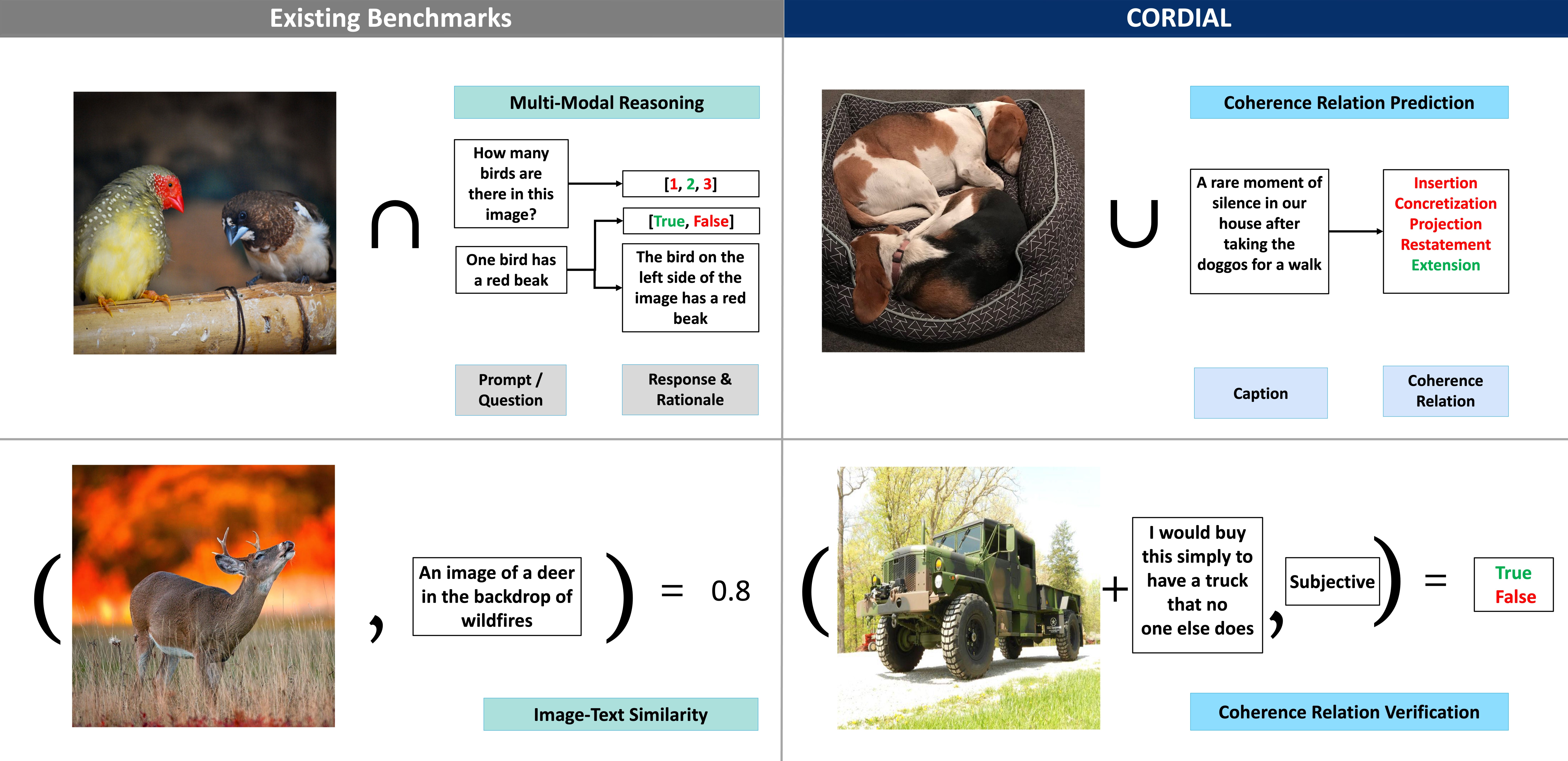"}}
    \caption{{\name} presents a combination of literal and pragmatic relations for analyzing the intermodal reasoning capabilities of MLLMs. We evaluate MLLMs on the task of Multimodal Discourse Analysis through the prediction and verification of Coherence Relations across three different discourse domains.}
    \label{fig:main_arch}
\end{figure*}

The recent advancements in Multimodal Large Language Models (MLLMs) enable them to effectively capture diverse representations of problem domains \cite{Alayrac2022-vq, Chen2024-vu, Pichai2024-xj, Liu2024-ay}. These MLLMs are capable of adapting to various downstream tasks with limited data through Parameter-Efficient Fine-Tuning (PEFT) \cite{Hu2021-ft} and In-Context Learning (ICL) \cite{Brown2020-hw} approaches. Existing Vision-based MLLM benchmarks assess different aspects of model performance such as Perception, Cognition, and Reasoning \cite{Li2024-bt} through various downstream tasks. 

Current benchmark design strategies often focus on evaluating the ability of MLLMs to utilize the intersection of input sources to solve a common problem \cite{Kruk2019-ac}. Although this helps assess the model's ability to interpret its inputs factually and logically, it does {\em not fully capture the model's understanding of the relationships between these modalities}. Similarly, benchmarks that evaluate the alignment between images and text \cite{Thrush2022-yf}, utilize curated or synthetically generated image-text pairs. These methods focus solely on literal relations that measure the level of overlap between the image and text. On the other hand, pragmatic cues provide information on non-literal relations where the true intent/message of an example may not be directly referenced in both modalities as shown in Figure \ref{fig:main_arch}. These cues are leveraged routinely in real-world multimodal discourses, which are characterized by the use of multiple modes of communication to convey different components of a message. Multimodal Discourse Analysis (MDA) studies how the interaction between these different modes can create semiotic meaning \cite{Kress2009-iy}.

To operationalize the assessment of these intermodal relationships, we turn to theories of {\em Discourse Coherence} \cite{Hobbs1978-em}, which offer a way to quantify the organization and flow of ideas across information sources. From these theories, we focus on the concept of {\em Coherence Relations} \cite{Alikhani2019-dz}, which provides a finite structure to link different parts of a discourse. Recent studies have extended these traditionally text-only theories to multimodal discourses, showing that Coherence Relations can be effectively applied to image-text pairs \cite{Alikhani2020-nr}. With Coherence Relations being a fundamental aspect of human communication, we evaluate whether MLLMs can effectively predict and verify these relations.

In this work, we propose the {\name} (\underline{CO}herence \underline{R}elations in \underline{D}iscourse for \underline{I}mages \underline{A}nd \underline{L}anguage), the first benchmark for evaluating MLLMs on the task of MDA. {\name} consists of a diverse set of Coherence Relations across three different discourse domains: Disaster Management, Social Media, and Online Articles. Each domain also offers different levels of complexity in the evaluated Coherence Relations, from binary relations to more challenging settings such as multi-class and multi-label relations assigned by human annotators. We evaluate the performance of 10+ MLLMs on {\name}, focusing on three research questions: 

{\bf
\begin{enumerate}[leftmargin=1cm, label=RQ\arabic*:, itemsep=-0.5ex]
    \item Can MLLMs predict Coherence Relations effectively?
    \item Can MLLMs verify Coherence Relations accurately?
    \item Can we teach MLLMs to understand Coherence Relations better?
\end{enumerate}
}

Our analysis reveals that both Coherence Relation prediction (RQ1) and verification (RQ2) are challenging tasks for MLLMs when these relations focus on pragmatic cues. Although larger MLLMs perform better than their smaller, open-source counterparts, traditional classifier baselines consistently outperform them across discourse domains. To summarize, our key takeaways are as follows:

\begin{itemize}[leftmargin=3.3mm]
    \item We propose {\name}, the first benchmark for evaluating MLLMs for Multi-modal Discourse Analysis (MDA) using Coherence Relations.
    \item Our experiments show that MLLMs struggle to predict and verify Coherence Relations, especially when these relations are more pragmatic.
    \item We demonstrate the need for coherence-aware fine-tuning approaches to improve intermodal reasoning capabilities of MLLMs.
\end{itemize}

\begin{table*}[!ht]
    \scriptsize
    \centering

    \scalebox{0.84}{
        \begin{tabularx}{\linewidth}{@{} cccccc @{}}
        \toprule
        \textbf{Dataset} & \multicolumn{5}{c}{\textbf{Examples}} \\
        \midrule        

        \multirow{4}{*}{DisREL} & \multicolumn{2}{c}{\makebox[0.15\textwidth]{\includegraphics[width=0.15\textwidth]{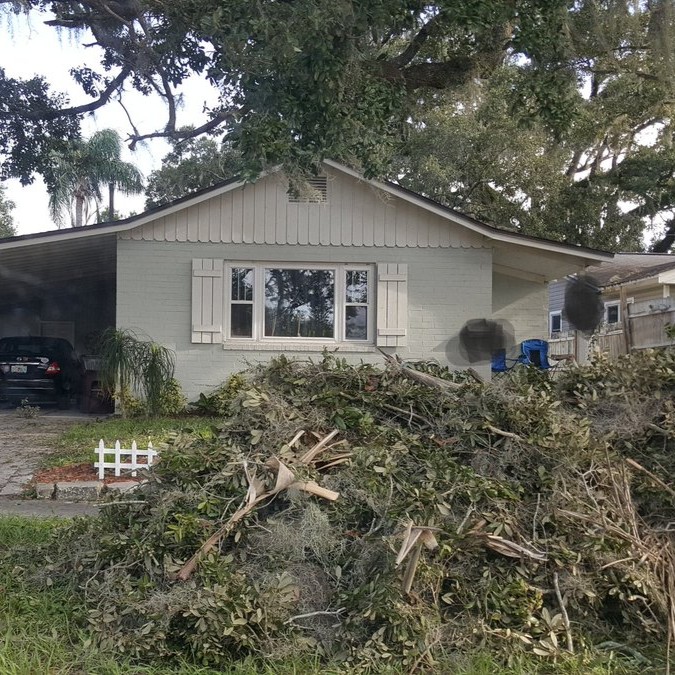}}} & & 
        \multicolumn{2}{c}{\makebox[0.15\textwidth]{\includegraphics[width=0.15\textwidth]{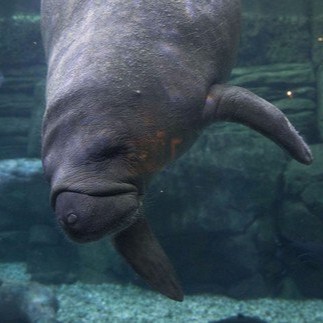}}} \\

        & \multicolumn{2}{c}{\multirow{2}{0.2\textwidth}{\centering Part of my {\color{red} pile of branches} after \#HurricaneIrma - still no power in \#Orlando}} & &
        \multicolumn{2}{c}{\multirow{2}{0.2\textwidth}{\centering Floridians rescue stranded {\color{red} manatees} as Irma sucks water from shores }} \\

        &  &  &  &  &  \\
        &  &  &  &  &  \vspace{0.1cm} \\

        & \multicolumn{2}{c}{\multirow{2}{0.15\textwidth}{\centering \textbf{Coherence Relation:} Similar}} & &
        \multicolumn{2}{c}{\multirow{2}{0.15\textwidth}{\centering \textbf{Coherence Relation:} Complementary}} \\
        &  &  &  &  &  \\
        
        \midrule
        
        \multirow{4}{*}{Tweet Subtitles} & \makebox[0.15\textwidth]{\includegraphics[width=0.15\textwidth]{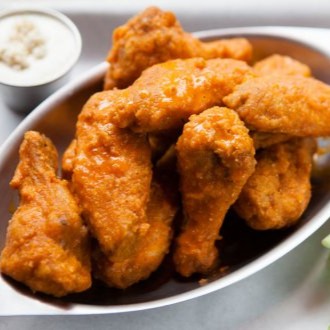}} & \makebox[0.15\textwidth]{\includegraphics[width=0.15\textwidth]{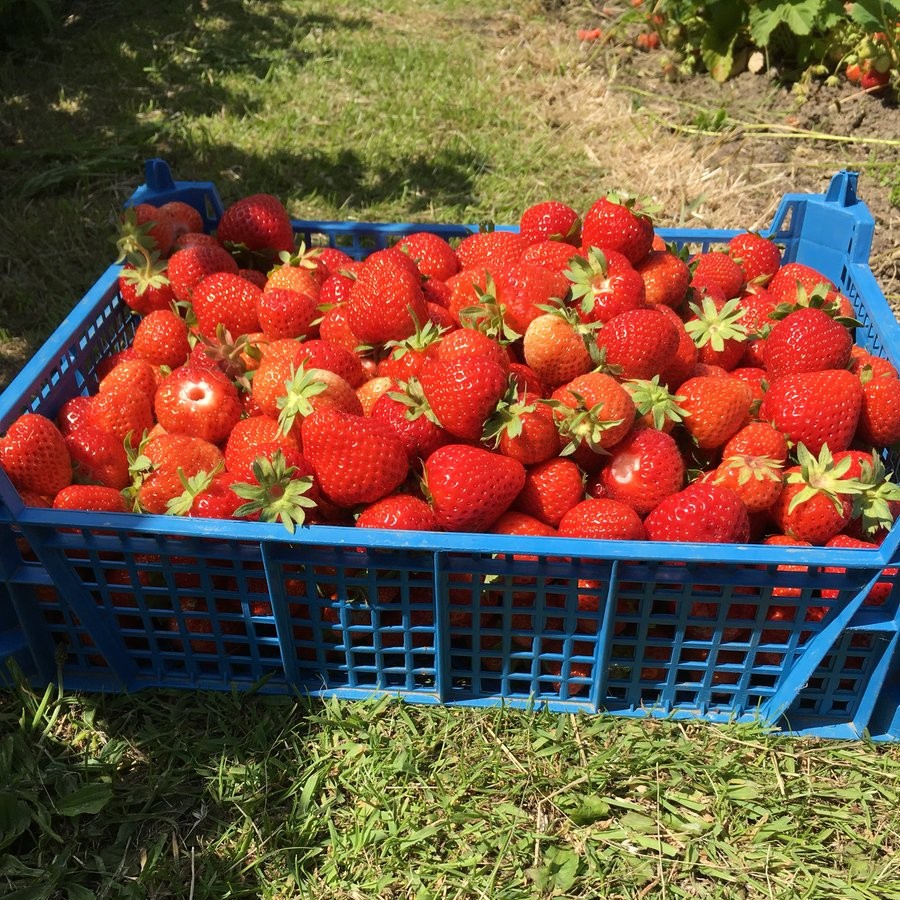}} & \makebox[0.15\textwidth]{\includegraphics[width=0.15\textwidth]{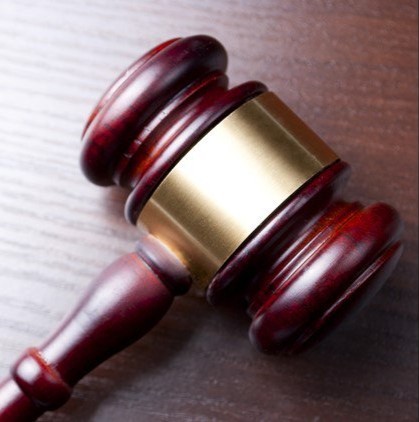}} & \makebox[0.15\textwidth]{\includegraphics[width=0.15\textwidth]{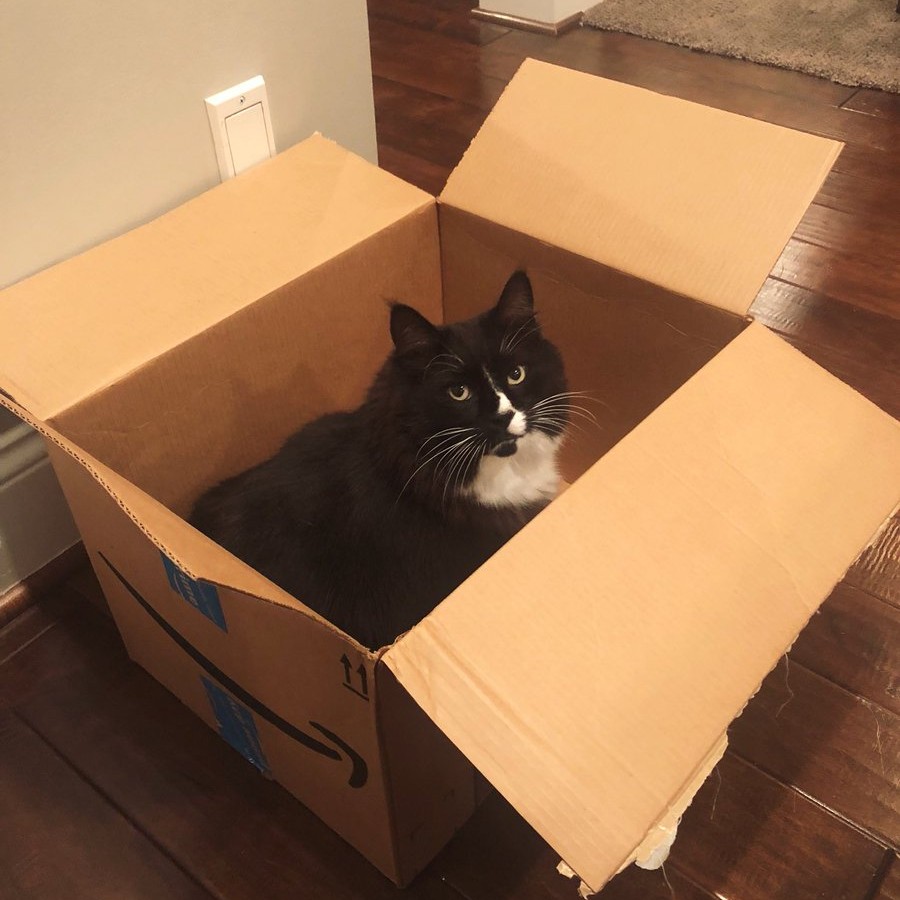}} & \makebox[0.15\textwidth]{\includegraphics[width=0.15\textwidth]{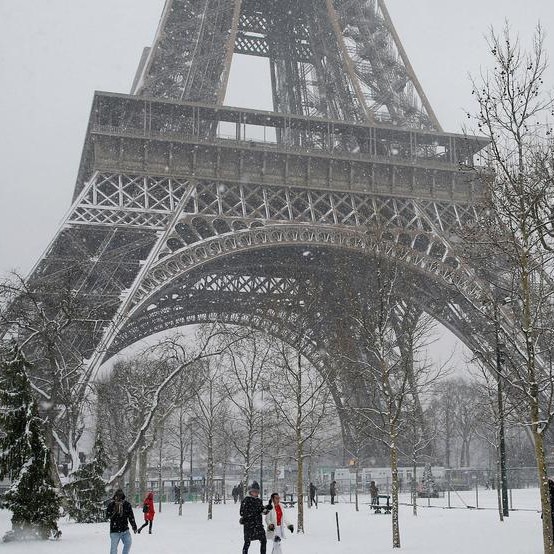}} \\
        
        & \multirow{3}{0.15\textwidth}{\centering Fresh never frozen {\color{red} jumbo wings} tossed in a housemade buffalo sauce. Yum!} & \multirow{3}{0.15\textwidth}{\centering Freshly picked off my allotment today, \\ well chuffed. \break {\color{orange} (strawberry)}} & \multirow{3}{0.15\textwidth}{\centering Cartel leader whose arrest sparked killings is sentenced to prison in Dallas {\color{red} court}} & \multirow{3}{0.15\textwidth}{\centering Amazon Prime delivers anything these days! {\color{orange} (delivering a cat)}} & \multirow{3}{0.15\textwidth}{\centering {\color{red} Eiffel Tower} shuts down as {\color{red} snow, freezing rain} pummel France} \\
        
        &  &  &  &  &  \\
        &  &  &  &  &  \\
        &  &  &  &  &  \vspace{0.1cm} \\

        & \multirow{2}{0.15\textwidth}{\centering \textbf{Coherence Relation:} Concretization} & \multirow{2}{0.15\textwidth}{\centering \textbf{Coherence Relation:} Insertion} & \multirow{2}{0.15\textwidth}{\centering \textbf{Coherence Relation:} Projection} & \multirow{2}{0.15\textwidth}{\centering \textbf{Coherence Relation:} Extension} & \multirow{2}{0.15\textwidth}{\centering \textbf{Coherence Relation:} Restatement} \\
        &  &  &  &  &  \\

        \midrule
        
        \multirow{4}{*}{CLUE} & \makebox[0.15\textwidth]{\includegraphics[width=0.15\textwidth]{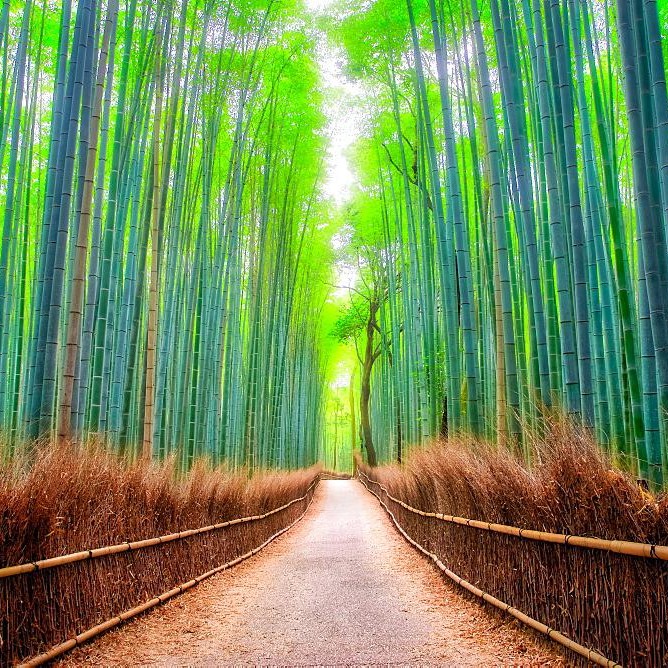}} & \makebox[0.15\textwidth]{\includegraphics[width=0.15\textwidth]{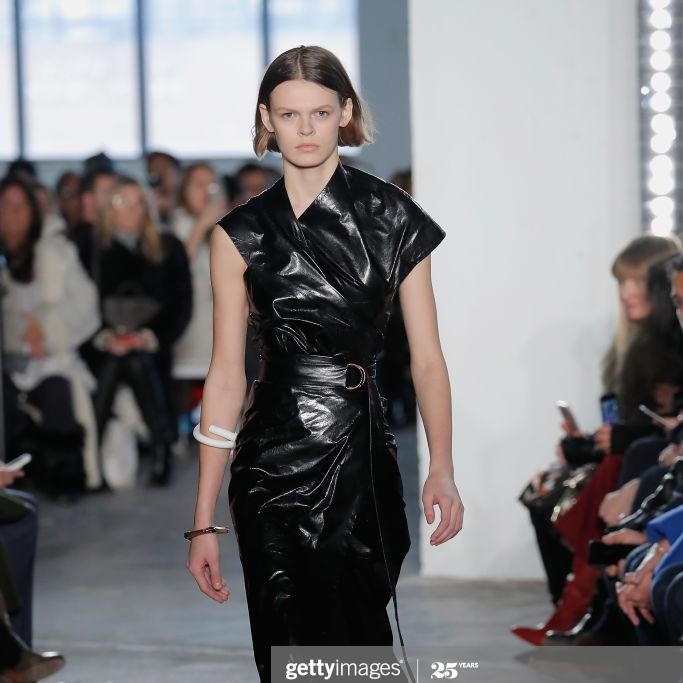}} & 
        \makebox[0.15\textwidth]{\includegraphics[width=0.15\textwidth]{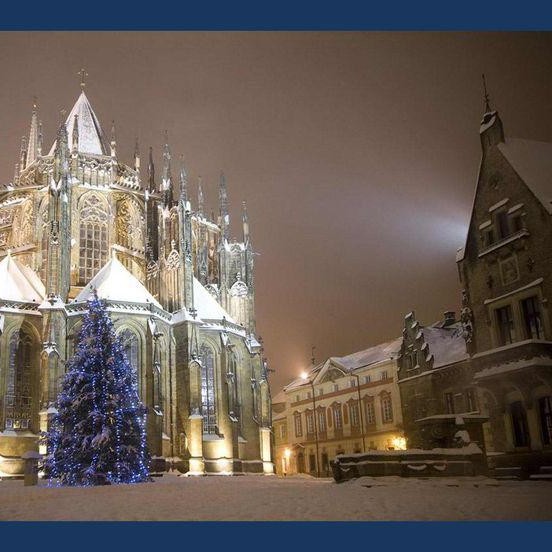}} & \makebox[0.15\textwidth]{\includegraphics[width=0.15\textwidth]{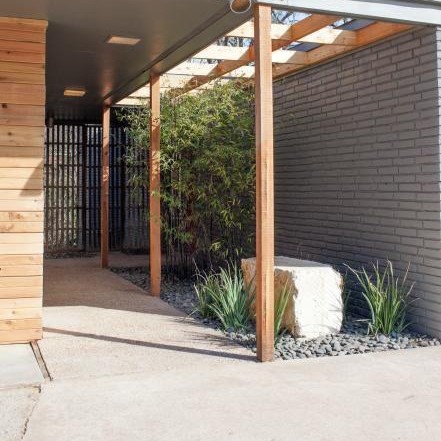}} & \makebox[0.15\textwidth]{\includegraphics[width=0.15\textwidth]{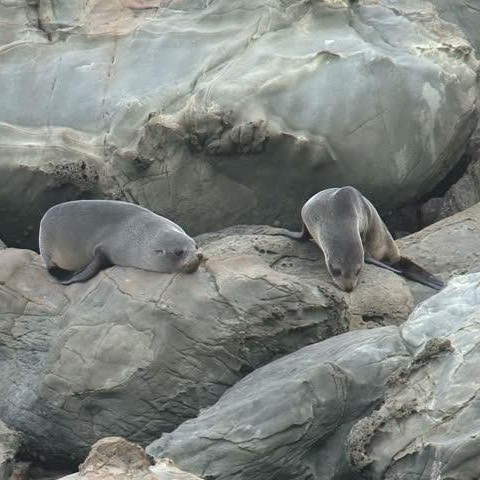}} \\
        
        & \multirow{3}{0.15\textwidth}{\centering A {\color{red} path} winds through an ancient {\color{red} bamboo forest}} & \multirow{3}{0.15\textwidth}{\centering A model {\color{red} walks} the runway for the collection during, {\color{red} fashion week}} & \multirow{3}{0.15\textwidth}{\centering A city in winter is such a {\color{red} beautiful} city} & \multirow{3}{0.15\textwidth}{\centering People know that {\color{red} curb appeal} is not a thing to take lightly when {\color{red} remodeling a home}} & \multirow{3}{0.15\textwidth}{\centering Seals {\color{red} fighting} for a spot to sleep on the rocks} \\
        
        &  &  &  &  &  \\
        &  &  &  &  &  \\
        &  &  &  &  & \vspace{0.1cm} \\

        & \multirow{2}{0.15\textwidth}{\centering \textbf{Coherence Relations:} {\color{blue} Visible}} & \multirow{2}{0.15\textwidth}{\centering \textbf{Coherence Relations:} Visible, {\color{blue} Meta}, Action} & \multirow{2}{0.15\textwidth}{\centering \textbf{Coherence Relations:} {\color{blue} Subjective}, Story} & \multirow{2}{0.15\textwidth}{\centering \textbf{Coherence Relations:} {\color{blue} Story}} & \multirow{2}{0.15\textwidth}{\centering \textbf{Coherence Relations:} {\color{blue} Action}} \\
        &  &  &  &  &  \\
        
        \bottomrule
        \end{tabularx}
    }
    \caption{Examples from each dataset for all Coherence Relations. The words in {\color{red} red} are important cues present in the caption, while the words in {\color{orange} orange} show pragmatic cues inferred from the image-text pair. The relations highlighted in {\color{blue} blue} are the selected relations for CLUE Single-Label.} 
    \label{table:examples-cr}
\end{table*}
\begin{figure*}
    \centering
    \includegraphics[width=0.85\linewidth]{"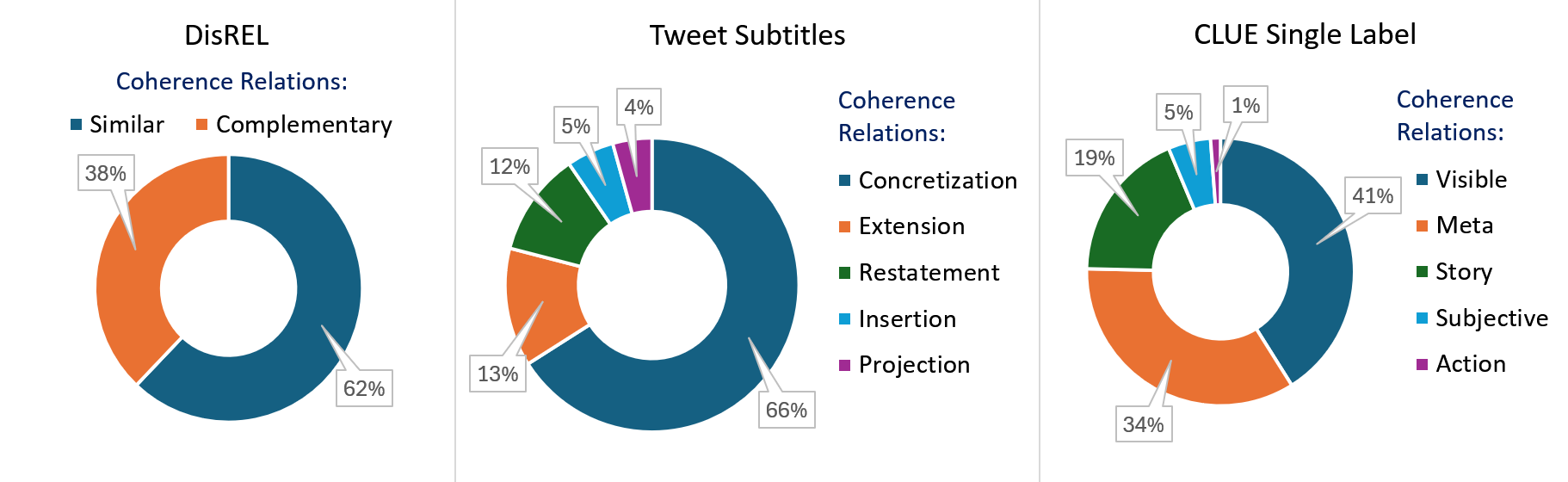"}
    \caption{An overview of the Image-Text label (i.e., Coherence Relations) distributions across {\name}}
    \label{fig:data_dist}
\end{figure*}

\section{Related Work}

\paragraph{Multimodal Large Language Models}
MLLMs are fundamentally generative models that combine Large Language Models (LLM) \cite{Brown2020-hw} with multimodal encoders \cite{Dosovitskiy2021-hj}. In recent years, several new MLLMs have been released, based on various proprietary \cite{OpenAI2024-hr, AnthropicUnknown-hu, Pichai2024-xj} and open-source LLM backbones \cite{Liu2023-os, Wu2024-ip, Bai2023-lu}. These models have shown impressive performance on a variety of downstream reasoning tasks, including Visual Question Answering \cite{Wu2024-pq}, Document Analysis \cite{Lv2023-ex}, Embodied AI agents \cite{Shek2024-dx}, etc.

\paragraph{MLLM Reasoning Benchmarks}
Recent works that have proposed benchmarks evaluating vision language reasoning, focus on assessing different facets of their input modalities. Visual Reasoning benchmarks measure the capability of these models to understand spatial and object-level relations among image components \cite{Kamath2023-ls, Rajabi2024-sa, Nie2024-nv, Thrush2022-yf, Kamoi2024-fc}. Contextual Reasoning benchmarks demonstrate how MLLMs interpret in-context examples and compositional language prompts \cite{Zong2024-xa, Wu2024-pq, Shao2024-jq, Zeng2024-sp}. Finally, Knowledge-based reasoning assesses how models recall knowledge from intrinsic and extrinsic sources to answer factual and logical questions \cite{Johnson2016-ut, Xenos2023-gz, Lu2022-aw}. Although these benchmarks measure how multimodal prompts can be efficiently understood to solve a candidate task, intermodal reasoning with real-world discourses has been less studied.

\paragraph{Image-Text Relationships}
Quantifying image-text relationships accurately has been an active area of research in the era of Vision Language Models (VLMs). Traditional VLMs translate images and text into a common representation space and compute the degree of similarity based on the distance between these embeddings \cite{Radford2021-ro, Jia2021-lq, Caron2021-cq, Hessel2021-we}. However, these methods failed to capture human preferences in image-text matching accurately across different task domain benchmarks \cite{Anantha-Ramakrishnan2024-sv, Ross2024-np, Anantha-Ramakrishnan2024-rm}. To include human feedback in the process of predicting similarity scores, content-based models trained on human-annotated similarity scores were introduced \cite{Wu2023-qy, Kirstain2023-km, Xu2023-rj}. Apart from similarity scores, taxonomies have been proposed to quantify different types of linkages between image-text pairs \cite{Marsh2003-hz, Vempala2019-lh, Kruk2019-ac, Bateman2014-dx}. In particular, multimodal coherence relations have been shown to sufficiently capture different aspects of image-text intents for various vision-language tasks \cite{Alikhani2019-kn, Inan2021-ip, Alikhani2023-nm, Alikhani2020-nr, Xu2022-ie}.
\section{The {\name} Benchmark}

\subsection{Motivation}
With Coherence Relations providing a finite representation of image-text linkages, we aim to measure MLLM performance through relation classification and verification tasks. Traditional alignment benchmarks often evaluate models using similarity scores. But multiple states of alignment between image-text pairs can exist, at the object-level, scene-level, or even at the discourse-level \cite{Xu2022-ie}. A pragmatic understanding of the context surrounding these pairs informs our ability to describe this alignment accurately. Thus, similarity scores alone may not be sufficient to capture the true performance of MLLMs. Additionally, with Coherence Relations being context-driven, the type of relations present in a discourse can vary across different domains. This necessitates the evaluation of MLLMs on multiple real-world discourse domains to assess their generalization capabilities. With MLLMs-as-a-judge \cite{Chen2024-cr} becoming more popular in tasks where acquiring human judgment is expensive and time-consuming, the importance of this task is further highlighted. We carefully pick and curate \textit{real-world image-text pairs} with \textit{expert human annotations} with the pre-processing details described in Appendix Section \ref{appendix-data-prep}. The three different discourse domains we evaluate are: Disaster Management, Social Media, and Online Articles.

\subsection{Coherence Relations}
Each dataset we include in {\name} assesses a unique set of Coherence Relations. To understand how communication in a discourse can be quantified by Coherence Relations, we turn to the Theory of Coherence \cite{Hobbs1978-em}. We define communication as the transfer of information and ideas from a speaker to a listener. For successful communication, a discourse needs to satisfy 4 conditions: (1) The message contents should be present in the discourse (2) The message must be relevant to the overall context of the discourse (3) Any new/unpredictable attributes of the message must build on the listener's existing world knowledge (4) The speaker must provide cues to guide the listener to graph their intended meaning. The goal of defining Coherence Relations is to serve any of the above-mentioned communicative functions. This way, for tasks such as MDA, we can analyze the communicative patterns present in a multimodal discourse. We consider Coherence Relations to be a constrained set of connections that describe the structural and causal relationships between different parts of a discourse. Consider the examples from Table \ref{table:examples-cr}, certain relations such as Visible and Concretization deal with presenting the same message content across modalities. On the other hand, relations such as Insertion and Extension require the reader to understand the union of information along with the context surrounding each modality to get the full message.

\subsection{Data Sources} \label{data-sources}
To construct our benchmark, we leverage existing datasets that provide image-text pairs along with human-annotated Coherence Relations across different discourse domains. We select three datasets that offer a diverse set of Coherence Relations: DisRel (Disaster Management), Tweet Subtitles (Social Media), and CLUE (Online Articles).

\paragraph{DisRel} 
This dataset \cite{Sosea2021-hr} explores the relationship of image-text pairs from disaster-related tweets, with labels collected through crowd-sourcing on Amazon MTurk. The dataset contains 4600 multimodal tweets with a test set size of \textit{500 examples} with a 50\% split between the two classes:

\begin{itemize}[leftmargin=3.3mm]
    \item \textbf{Similar}: The image and text share the same focus and attempt to convey the same message. There exists a significant overlap in the information conveyed between modalities. 
    \item \textbf{Complementary}:  The image and text do not share the same focus, but one modality helps understand the other better. Both modalities provide independent information which when combined, provide a more complete picture of the message/event. There may be divergence in the information conveyed between modalities.

\end{itemize}

\paragraph{Tweet Subtitles}
To measure cross-modal coherence relations between image and text, this dataset \cite{Xu2022-ie} contains 16000 image-text pairs sourced from Twitter on open-domain topics. The test set for this dataset consists of \textit{1600 examples}, which is 10\% of the entire dataset. The dataset provides single-label annotations from expert annotators on 3 entity-level and 2 scene-level relations:

\begin{itemize}[leftmargin=3.3mm]
    \item \textbf{Insertion (Entity-level)}: Both the text and the image focus on the same visual entity but it is not explicitly mentioned in the text.
    \item \textbf{Concretization (Entity-level)}: Both the text and image contain a mention of the main visual entity but may differ in types of details shared.
    \item \textbf{Projection (Entity-level)}: The main entity mentioned in the text is implicitly related to the visual objects present in the image. The image contains a reference to objects related to the main entity rather than the entity itself.
    \item \textbf{Restatement (Scene-level)}: The text directly describes the image contents. Both modalities convey the same message.
    \item \textbf{Extension (Scene-level)}: The image expands upon the story or idea in the text, presenting new elements or elaborations, effectively filling in narrative gaps left by the text.
\end{itemize}

\paragraph{CLUE} \label{clue-labels}
This dataset presents a novel conceptualization of image-text relations by extending text-only coherence relations to the multimodal setting \cite{Alikhani2020-nr}. The publicly available version of the dataset contains 4770 image-text pairs sourced from the Conceptual Captions Dataset \cite{Sharma2018-tr}. The samples were provided multi-label annotations by expert annotators for 5 different relationship types:

\begin{itemize}[leftmargin=3.3mm]
    \item \textbf{Visible}: The text presents information that is intended to recognizably characterize what is depicted in the image.
    \item \textbf{Action}: The text describes an extended, dynamic process in which the moment captured in the image is a representative snapshot.
    \item \textbf{Meta}: The text allows the reader to draw inferences not just about the scene depicted in the image but about the production and presentation of the image itself.
    \item \textbf{Subjective}: The text provides information about the speaker's reaction to, or evaluation of, what is depicted in the image.
    \item \textbf{Story}: The text provides a freestanding description of the circumstances depicted in the image, analogous to including instructional, explanatory, and other background relations.
\end{itemize}

We evaluate this dataset in two different settings: Multi-Label (ML) and Single-Label (SL). In the ML setting, we treat the dataset as a multi-label classification task where MLLMs predict all applicable labels. For CLUE SL, we follow the original dataset's label mapping strategy to select the most applicable label from the present annotations for each sample \cite{Alikhani2020-nr}. This provides two different settings for evaluating MLLM's understanding of coherence relations on the same image-text pairs with \textit{1183 examples} in the test set. 

\subsection{Baseline Classifier} \label{classifier}
Our goal of including a baseline classifier is to capture the existing signal in our datasets and to provide a reference point for MLLM performance. Understanding that human annotations can be noisy, we utilize this simple, generalizable classifier to identify relations where MLLMs are particularly under-performing on our benchmark. We employ CLIP Text and Image encoders to extract multimodal embeddings in a zero-shot manner \cite{Radford2021-ro}. We then train a Multi-Layer Perceptron (MLP) classifier using these embeddings on the train sets of each of these datasets to predict Coherence Relations. This ensures that our classifier is not biased towards any specific domain and can generalize across different discourse contexts. More details about the classifier are present in Appendix Section \ref{apendix-classifier}.
\begin{figure}
    \centering
    \includegraphics[width=0.9\linewidth]{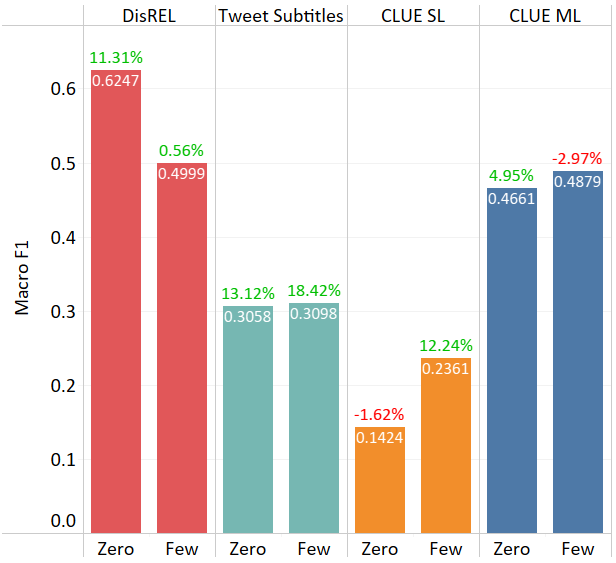}
    \caption{\% Loss/Gain after fine-tuning Llama 3.2-V. Fine-tuning shows significant performance gains, either on zero-shot or few-shot prompts across all 4 settings}
    \label{fig:finetuning}
\end{figure}

\begin{table}[t]
    \centering
    \scriptsize
        \begin{tabularx}{\columnwidth}{@{} l | Y|Y @{} Y| l @{}}
        \toprule

        \textbf{Model} & \textbf{Prompt} & \textbf{Sim} & \textbf{Compl} & \textbf{Macro F1} \\
        \midrule

        Random Guess & Baseline & 0.490 & 0.478 & 0.484 \\
        \midrule
        \midrule
        \multirow{2}{*}{LLaVA 1.6 7B} & Zero & 0.253 & 0.541 & 0.397 \\
        & CoT & 0.544 & 0.489 & 0.516  \tikz[baseline=(X.base)]{\node[fill=green!50, rounded corners=2pt, inner sep=0pt] (X) {\strut \tiny ↑30.0\%}; } \\
        \midrule

        \multirow{2}{*}{LLaVA 1.6 13B} & Zero & 0.666 & 0.000 & 0.333 \\
        & CoT & 0.408 & 0.675 & 0.542  \tikz[baseline=(X.base)]{\node[fill=green!50, rounded corners=2pt, inner sep=0pt] (X) {\strut \tiny ↑62.8\%}; } \\
        \midrule

        \multirow{3}{*}{LLaVA 1.6 34B} & Zero & 0.000 & 0.666 & 0.333 \\
        & Few & 0.139 & 0.679 & 0.409  \tikz[baseline=(X.base)]{\node[fill=green!50, rounded corners=2pt, inner sep=0pt] (X) {\strut \tiny ↑22.8\%}; } \\
        & CoT & 0.353 & 0.571 & 0.462  \tikz[baseline=(X.base)]{\node[fill=green!50, rounded corners=2pt, inner sep=0pt] (X) {\strut \tiny ↑38.7\%}; } \\
        \midrule

        \multirow{3}{*}{LLaVA OneVision 7B} & Zero & 0.626 & 0.391 & 0.509 \\
        & Few & 0.549 & 0.541 & 0.545  \tikz[baseline=(X.base)]{\node[fill=green!50, rounded corners=2pt, inner sep=0pt] (X) {\strut \tiny ↑7.1\%}; } \\
        & CoT & 0.549 & 0.601 & 0.575  \tikz[baseline=(X.base)]{\node[fill=green!50, rounded corners=2pt, inner sep=0pt] (X) {\strut \tiny ↑13.0\%}; } \\
        \midrule

        \multirow{3}{*}{Qwen2-VL 7B} & Zero & 0.654 & 0.268 & 0.461 \\
        & Few & 0.664 & 0.148 & 0.406  \tikz[baseline=(X.base)]{\node[fill=red!50, rounded corners=2pt, inner sep=0pt] (X) {\strut \tiny ↓11.9\%}; } \\
        & CoT & 0.446 & 0.602 & 0.524  \tikz[baseline=(X.base)]{\node[fill=green!50, rounded corners=2pt, inner sep=0pt] (X) {\strut \tiny ↑13.7\%}; } \\
        \midrule

        \multirow{3}{*}{Llama 3.2 Vision 11B} & Zero & 0.388 & 0.635 & 0.512 \\
        & Few & 0.509 & 0.479 & 0.494  \tikz[baseline=(X.base)]{\node[fill=red!50, rounded corners=2pt, inner sep=0pt] (X) {\strut \tiny ↓3.5\%}; } \\
        & CoT & 0.292 & 0.615 & 0.453  \tikz[baseline=(X.base)]{\node[fill=red!50, rounded corners=2pt, inner sep=0pt] (X) {\strut \tiny ↓11.5\%}; } \\
        \midrule

        \multirow{3}{*}{Phi3.5 Vision 4.2B} & Zero & 0.655 & 0.177 & 0.416 \\
        & Few & 0.409 & 0.662 & 0.536  \tikz[baseline=(X.base)]{\node[fill=green!50, rounded corners=2pt, inner sep=0pt] (X) {\strut \tiny ↑28.8\%}; } \\
        & CoT & 0.549 & 0.601 & 0.575  \tikz[baseline=(X.base)]{\node[fill=green!50, rounded corners=2pt, inner sep=0pt] (X) {\strut \tiny ↑38.2\%}; } \\
        \midrule

        \multirow{3}{*}{InternVL 2.5 26B} & Zero & 0.618 & 0.698 & 0.658 \\
        & Few & 0.633 & 0.633 & 0.633  \tikz[baseline=(X.base)]{\node[fill=red!50, rounded corners=2pt, inner sep=0pt] (X) {\strut \tiny ↓3.8\%}; } \\
        & CoT & 0.393 & 0.670 & 0.531  \tikz[baseline=(X.base)]{\node[fill=red!50, rounded corners=2pt, inner sep=0pt] (X) {\strut \tiny ↓19.3\%}; } \\
        \midrule 
        \midrule

        \multirow{3}{*}{GPT-4o} & Zero & 0.025 & 0.667 & 0.346 \\
        & Few & 0.443 & 0.667 & 0.555  \tikz[baseline=(X.base)]{\node[fill=green!50, rounded corners=2pt, inner sep=0pt] (X) {\strut \tiny ↑60.4\%}; } \\
        & CoT & 0.361 & 0.676 & 0.519  \tikz[baseline=(X.base)]{\node[fill=green!50, rounded corners=2pt, inner sep=0pt] (X) {\strut \tiny ↑50.0\%}; } \\
        \midrule

        \multirow{3}{*}{Gemini 1.5 Flash} & Zero & 0.714 & 0.715 & \underline{0.715} \\
        & Few & 0.363 & 0.688 & 0.525  \tikz[baseline=(X.base)]{\node[fill=red!50, rounded corners=2pt, inner sep=0pt] (X) {\strut \tiny ↓26.6\%}; } \\
        & CoT & 0.593 & 0.699 & 0.646  \tikz[baseline=(X.base)]{\node[fill=red!50, rounded corners=2pt, inner sep=0pt] (X) {\strut \tiny ↓9.7\%}; } \\
        \midrule

        \multirow{3}{*}{Gemini 1.5 Pro} & Zero & 0.719 & 0.679 & 0.699 \\
        & Few & 0.611 & \textbf{0.727} & 0.669  \tikz[baseline=(X.base)]{\node[fill=red!50, rounded corners=2pt, inner sep=0pt] (X) {\strut \tiny ↓4.3\%}; } \\
        & CoT & 0.630 & \underline{0.717} & 0.673  \tikz[baseline=(X.base)]{\node[fill=red!50, rounded corners=2pt, inner sep=0pt] (X) {\strut \tiny ↓3.7\%}; } \\
        \midrule

        \multirow{3}{*}{Claude 3.5 Sonnet v2} & Zero & \underline{0.722} & 0.615 & 0.669 \\
        & Few & 0.710 & 0.559 & 0.634  \tikz[baseline=(X.base)]{\node[fill=red!50, rounded corners=2pt, inner sep=0pt] (X) {\strut \tiny ↓5.2\%}; } \\
        & CoT & 0.603 & 0.703 & 0.653  \tikz[baseline=(X.base)]{\node[fill=red!50, rounded corners=2pt, inner sep=0pt] (X) {\strut \tiny ↓2.4\%}; } \\
        \midrule 
        \midrule

        CLIP Classifier & Baseline & \textbf{0.750} & 0.715 & \textbf{0.733} \\

        \bottomrule

        \end{tabularx}
       \caption{Results for Coherence Relation Prediction on DisRel. The coherence relations predicted are Similar (Sim) and Complementary (Compl).}
        \label{table:metrics_disrel}
\end{table}

\begin{table}[!ht]
    \centering
    \scalebox{0.50}{
        \begin{tabularx}{1.97\linewidth}{@{} l|Y| Y @{} Y @{} Y @{} Y @{} Y| l @{}}
        \toprule
        \textbf{Model} & \textbf{Prompt} & \textbf{Ins} & \textbf{Concr} & \textbf{Proj} & \textbf{Restmt} & \textbf{Ext} & \textbf{Macro F1} \\
        \midrule

        Random Guess & Baseline & 0.094 & 0.340 & 0.068 & 0.123 & 0.165 & 0.158 \\
        \midrule
        \midrule
        \multirow{2}{*}{LLaVA 1.6 7B} & Zero & 0.000 & 0.693 & 0.062 & 0.066 & 0.082 & 0.181 \\
        & CoT & 0.019 & 0.822 & 0.081 & 0.050 & 0.114 & 0.217  \tikz[baseline=(X.base)]{\node[fill=green!50, rounded corners=2pt, inner sep=0pt] (X) {\strut \small ↑19.9\%}; } \\
        \midrule

        \multirow{2}{*}{LLaVA 1.6 13B} & Zero & 0.085 & 0.044 & 0.000 & 0.000 & 0.095 & 0.045 \\
        & CoT & 0.070 & 0.477 & 0.000 & 0.122 & 0.054 & 0.145  \tikz[baseline=(X.base)]{\node[fill=green!50, rounded corners=2pt, inner sep=0pt] (X) {\strut \small ↑222.2\%}; } \\
        \midrule

        \multirow{3}{*}{LLaVA 1.6 34B} & Zero & 0.000 & 0.176 & 0.094 & 0.104 & 0.253 & 0.125 \\
        & Few & 0.026 & 0.630 & 0.198 & 0.060 & 0.211 & 0.225  \tikz[baseline=(X.base)]{\node[fill=green!50, rounded corners=2pt, inner sep=0pt] (X) {\strut \small ↑80.0\%}; } \\
        & CoT & 0.024 & 0.063 & 0.108 & 0.154 & 0.169 & 0.104  \tikz[baseline=(X.base)]{\node[fill=red!50, rounded corners=2pt, inner sep=0pt] (X) {\strut \small ↓16.8\%}; } \\
        \midrule

        \multirow{3}{*}{LLaVA OneVision 7B} & Zero & 0.023 & 0.000 & 0.066 & 0.125 & 0.032 & 0.049 \\
        & Few & 0.067 & 0.000 & 0.087 & 0.071 & 0.177 & 0.081  \tikz[baseline=(X.base)]{\node[fill=green!50, rounded corners=2pt, inner sep=0pt] (X) {\strut \small ↑65.3\%}; } \\
        & CoT & 0.062 & 0.005 & 0.057 & 0.124 & 0.101 & 0.070  \tikz[baseline=(X.base)]{\node[fill=green!50, rounded corners=2pt, inner sep=0pt] (X) {\strut \small ↑42.9\%}; } \\
        \midrule

        \multirow{3}{*}{Qwen2-VL 7B} & Zero & 0.000 & 0.728 & 0.121 & 0.142 & 0.011 & 0.201 \\
        & Few & 0.094 & 0.148 & 0.078 & 0.144 & 0.068 & 0.106  \tikz[baseline=(X.base)]{\node[fill=red!50, rounded corners=2pt, inner sep=0pt] (X) {\strut \small ↓47.3\%}; } \\
        & CoT & 0.156 & 0.167 & 0.068 & 0.170 & 0.000 & 0.112  \tikz[baseline=(X.base)]{\node[fill=red!50, rounded corners=2pt, inner sep=0pt] (X) {\strut \small ↓44.3\%}; } \\
        \midrule

        \multirow{3}{*}{Llama 3.2 Vision 11B} & Zero & 0.000 & 0.779 & 0.000 & 0.093 & 0.000 & 0.175 \\
        & Few & 0.035 & 0.388 & 0.000 & 0.092 & 0.113 & 0.126  \tikz[baseline=(X.base)]{\node[fill=red!50, rounded corners=2pt, inner sep=0pt] (X) {\strut \small ↓28.0\%}; } \\
        & CoT & 0.097 & 0.421 & 0.055 & 0.167 & 0.086 & 0.165  \tikz[baseline=(X.base)]{\node[fill=red!50, rounded corners=2pt, inner sep=0pt] (X) {\strut \small ↓5.7\%}; } \\
        \midrule

        \multirow{3}{*}{Phi3.5 Vision 4.2B} & Zero & 0.043 & \underline{0.790} & 0.109 & 0.171 & 0.030 & 0.229 \\
        & Few & 0.183 & 0.179 & 0.000 & 0.159 & 0.093 & 0.123  \tikz[baseline=(X.base)]{\node[fill=red!50, rounded corners=2pt, inner sep=0pt] (X) {\strut \small ↓46.3\%}; } \\
        & CoT & 0.025 & 0.745 & 0.164 & 0.156 & 0.022 & 0.223  \tikz[baseline=(X.base)]{\node[fill=red!50, rounded corners=2pt, inner sep=0pt] (X) {\strut \small ↓2.6\%}; } \\
        \midrule

        \multirow{3}{*}{InternVL 2.5 26B} & Zero & 0.101 & 0.389 & 0.090 & 0.090 & 0.011 & 0.136 \\
        & Few & 0.090 & 0.002 & 0.041 & 0.292 & 0.000 & 0.085  \tikz[baseline=(X.base)]{\node[fill=red!50, rounded corners=2pt, inner sep=0pt] (X) {\strut \small ↓37.5\%}; } \\
        & CoT & 0.118 & 0.450 & 0.102 & 0.199 & 0.083 & 0.190  \tikz[baseline=(X.base)]{\node[fill=green!50, rounded corners=2pt, inner sep=0pt] (X) {\strut \small ↑39.7\%}; } \\
        \midrule 
        \midrule

        \multirow{3}{*}{GPT-4o} & Zero & 0.126 & 0.564 & 0.111 & 0.200 & 0.167 & 0.234 \\
        & Few & 0.171 & 0.599 & 0.131 & 0.268 & 0.199 & 0.274  \tikz[baseline=(X.base)]{\node[fill=green!50, rounded corners=2pt, inner sep=0pt] (X) {\strut \small ↑17.1\%}; } \\
        & CoT & 0.076 & 0.346 & 0.146 & 0.217 & 0.187 & 0.194  \tikz[baseline=(X.base)]{\node[fill=red!50, rounded corners=2pt, inner sep=0pt] (X) {\strut \small ↓17.1\%}; } \\
        \midrule

        \multirow{3}{*}{Gemini 1.5 Flash} & Zero & 0.172 & 0.783 & 0.138 & 0.183 & 0.011 & 0.257 \\
        & Few & 0.027 & 0.681 & 0.139 & 0.257 & 0.193 & 0.259  \tikz[baseline=(X.base)]{\node[fill=green!50, rounded corners=2pt, inner sep=0pt] (X) {\strut \small ↑0.8\%}; } \\
        & CoT & 0.068 & 0.734 & 0.133 & 0.259 & 0.071 & 0.253  \tikz[baseline=(X.base)]{\node[fill=red!50, rounded corners=2pt, inner sep=0pt] (X) {\strut \small ↓1.6\%}; } \\
        \midrule

        \multirow{3}{*}{Gemini 1.5 Pro} & Zero & \underline{0.200} & 0.692 & 0.141 & 0.290 & 0.034 & 0.271 \\
        & Few & 0.113 & 0.661 & \underline{0.247} & 0.270 & 0.000 & 0.258  \tikz[baseline=(X.base)]{\node[fill=red!50, rounded corners=2pt, inner sep=0pt] (X) {\strut \small ↓4.8\%}; } \\
        & CoT & 0.102 & 0.657 & 0.101 & 0.278 & 0.022 & 0.232  \tikz[baseline=(X.base)]{\node[fill=red!50, rounded corners=2pt, inner sep=0pt] (X) {\strut \small ↓14.4\%}; } \\
        \midrule

        \multirow{3}{*}{Claude 3.5 Sonnet v2} & Zero & 0.132 & 0.764 & 0.183 & \underline{0.328} & 0.175 & 0.316 \\
        & Few & 0.144 & 0.567 & 0.122 & 0.285 & 0.246 & 0.273  \tikz[baseline=(X.base)]{\node[fill=red!50, rounded corners=2pt, inner sep=0pt] (X) {\strut \small ↓13.6\%}; } \\
        & CoT & 0.180 & 0.725 & 0.138 & 0.316 & \underline{0.256} & \underline{0.323}  \tikz[baseline=(X.base)]{\node[fill=green!50, rounded corners=2pt, inner sep=0pt] (X) {\strut \small ↑2.2\%}; } \\
        \midrule 
        \midrule

        CLIP Classifier & Baseline & \textbf{0.542} & \textbf{0.866} & \textbf{0.286} & \textbf{0.388} & \textbf{0.514} & \textbf{0.519} \\

        \bottomrule
        \end{tabularx}
        }
       \caption{Results for Coherence Relation Prediction on Tweet Subtitles. The Coherence Relations predicted are Insertion (Ins), Concretization (Concr), Projection (Proj), Restatement (Restmt) and Extension (Ext).}
        \label{table:metrics_tweets}
\end{table}

\begin{table}[!ht]
    \centering
    \scalebox{0.50}{
        \begin{tabularx}{1.97\linewidth}{@{} l|Y| Y @{} Y @{} Y @{} Y @{} Y| l @{}}
        \toprule
        \textbf{Model} & \textbf{Prompt} & \textbf{Visible} & \textbf{Subj} & \textbf{Action} & \textbf{Story} & \textbf{Meta} & \textbf{Macro F1} \\
        \midrule

        Random Guess & Baseline & 0.233 & 0.069 & 0.030 & 0.162 & 0.266 & 0.152 \\

        \midrule
        \midrule
        \multirow{2}{*}{LLaVA 1.6 7B} & Zero & 0.484 & 0.135 & 0.000 & 0.158 & 0.096 & 0.174 \\
        & CoT & 0.534 & 0.198 & 0.068 & 0.043 & 0.004 & 0.169  \tikz[baseline=(X.base)]{\node[fill=red!50, rounded corners=2pt, inner sep=0pt] (X) {\strut \small ↓2.9\%}; } \\
        \midrule

        \multirow{2}{*}{LLaVA 1.6 13B} & Zero & 0.541 & 0.027 & 0.039 & 0.158 & 0.000 & 0.153 \\
        & CoT & 0.529 & 0.043 & 0.054 & 0.034 & 0.016 & 0.135  \tikz[baseline=(X.base)]{\node[fill=red!50, rounded corners=2pt, inner sep=0pt] (X) {\strut \small ↓11.8\%}; } \\
        \midrule

        \multirow{3}{*}{LLaVA 1.6 34B} & Zero & 0.545 & 0.000 & 0.000 & 0.012 & 0.004 & 0.112 \\
        & Few & 0.457 & 0.097 & 0.058 & 0.318 & 0.086 & 0.203  \tikz[baseline=(X.base)]{\node[fill=green!50, rounded corners=2pt, inner sep=0pt] (X) {\strut \small ↑81.3\%}; } \\
        & CoT & 0.537 & 0.143 & 0.062 & 0.210 & 0.004 & 0.191  \tikz[baseline=(X.base)]{\node[fill=green!50, rounded corners=2pt, inner sep=0pt] (X) {\strut \small ↑70.5\%}; } \\
        \midrule

        \multirow{3}{*}{LLaVA OneVision 7B} & Zero & 0.541 & 0.000 & 0.087 & 0.043 & 0.000 & 0.134 \\
        & Few & 0.146 & 0.000 & 0.025 & 0.172 & 0.243 & 0.117  \tikz[baseline=(X.base)]{\node[fill=red!50, rounded corners=2pt, inner sep=0pt] (X) {\strut \small ↓12.7\%}; } \\
        & CoT & 0.535 & 0.000 & 0.048 & 0.092 & 0.000 & 0.135  \tikz[baseline=(X.base)]{\node[fill=green!50, rounded corners=2pt, inner sep=0pt] (X) {\strut \small ↑0.7\%}; } \\
        \midrule

        \multirow{3}{*}{Qwen2-VL 7B} & Zero & 0.533 & 0.068 & 0.000 & 0.034 & 0.000 & 0.127 \\
        & Few & 0.539 & 0.000 & 0.000 & 0.000 & 0.004 & 0.109  \tikz[baseline=(X.base)]{\node[fill=red!50, rounded corners=2pt, inner sep=0pt] (X) {\strut \small ↓14.2\%}; } \\
        & CoT & 0.530 & 0.156 & 0.057 & 0.080 & 0.004 & 0.166  \tikz[baseline=(X.base)]{\node[fill=green!50, rounded corners=2pt, inner sep=0pt] (X) {\strut \small ↑30.7\%}; } \\
        \midrule

        \multirow{3}{*}{Llama 3.2 Vision 11B} & Zero & 0.537 & 0.136 & \underline{0.098} & 0.023 & 0.000 & 0.159 \\
        & Few & 0.542 & 0.000 & 0.026 & 0.000 & 0.000 & 0.114  \tikz[baseline=(X.base)]{\node[fill=red!50, rounded corners=2pt, inner sep=0pt] (X) {\strut \small ↓28.3\%}; } \\
        & CoT & 0.533 & 0.189 & 0.026 & 0.083 & 0.020 & 0.170  \tikz[baseline=(X.base)]{\node[fill=green!50, rounded corners=2pt, inner sep=0pt] (X) {\strut \small ↑6.9\%}; } \\
        \midrule

        \multirow{3}{*}{Phi3.5 Vision 4.2B} & Zero & 0.542 & 0.038 & 0.053 & 0.104 & 0.000 & 0.147 \\
        & Few & 0.485 & 0.256 & 0.021 & 0.255 & 0.162 & 0.236  \tikz[baseline=(X.base)]{\node[fill=green!50, rounded corners=2pt, inner sep=0pt] (X) {\strut \small ↑60.5\%}; } \\
        & CoT & 0.534 & 0.000 & 0.087 & 0.083 & 0.000 & 0.141  \tikz[baseline=(X.base)]{\node[fill=red!50, rounded corners=2pt, inner sep=0pt] (X) {\strut \small ↓4.1\%}; } \\
        \midrule

        \multirow{3}{*}{InternVL 2.5 26B} & Zero & 0.558 & 0.273 & 0.071 & 0.312 & 0.027 & 0.248 \\
        & Few & 0.498 & 0.211 & 0.048 & 0.253 & 0.127 & 0.228  \tikz[baseline=(X.base)]{\node[fill=red!50, rounded corners=2pt, inner sep=0pt] (X) {\strut \small ↓8.1\%}; } \\
        & CoT & 0.537 & 0.333 & 0.052 & 0.254 & 0.087 & 0.252  \tikz[baseline=(X.base)]{\node[fill=green!50, rounded corners=2pt, inner sep=0pt] (X) {\strut \small ↑1.6\%}; } \\
        \midrule 
        \midrule

        \multirow{3}{*}{GPT-4o} & Zero & 0.544 & 0.345 & 0.064 & 0.178 & 0.065 & 0.239 \\
        & Few & 0.549 & 0.352 & 0.023 & 0.390 & 0.134 & 0.289  \tikz[baseline=(X.base)]{\node[fill=green!50, rounded corners=2pt, inner sep=0pt] (X) {\strut \small ↑20.9\%}; } \\
        & CoT & 0.558 & 0.321 & 0.054 & 0.324 & 0.024 & 0.256  \tikz[baseline=(X.base)]{\node[fill=green!50, rounded corners=2pt, inner sep=0pt] (X) {\strut \small ↑7.1\%}; } \\
        \midrule

        \multirow{3}{*}{Gemini 1.5 Flash} & Zero & 0.543 & 0.215 & 0.091 & 0.168 & 0.020 & 0.207 \\
        & Few & 0.543 & 0.380 & 0.054 & 0.402 & 0.071 & 0.290  \tikz[baseline=(X.base)]{\node[fill=green!50, rounded corners=2pt, inner sep=0pt] (X) {\strut \small ↑40.1\%}; } \\
        & CoT & 0.557 & 0.300 & 0.000 & 0.329 & 0.072 & 0.252  \tikz[baseline=(X.base)]{\node[fill=green!50, rounded corners=2pt, inner sep=0pt] (X) {\strut \small ↑21.7\%}; } \\
        \midrule

        \multirow{3}{*}{Gemini 1.5 Pro} & Zero & \textbf{0.559} & 0.329 & 0.039 & 0.440 & 0.112 & 0.296 \\
        & Few & 0.531 & 0.391 & 0.070 & \underline{0.451} & 0.253 & 0.339  \tikz[baseline=(X.base)]{\node[fill=green!50, rounded corners=2pt, inner sep=0pt] (X) {\strut \small ↑14.5\%}; } \\
        & CoT & \underline{0.558} & 0.330 & 0.000 & 0.350 & 0.057 & 0.259  \tikz[baseline=(X.base)]{\node[fill=red!50, rounded corners=2pt, inner sep=0pt] (X) {\strut \small ↓12.5\%}; } \\
        \midrule

        \multirow{3}{*}{Claude 3.5 Sonnet v2} & Zero & 0.516 & \underline{0.408} & 0.070 & 0.439 & 0.113 & 0.309 \\
        & Few & 0.467 & \textbf{0.430} & 0.077 & 0.434 & \underline{0.338} & \underline{0.349}  \tikz[baseline=(X.base)]{\node[fill=green!50, rounded corners=2pt, inner sep=0pt] (X) {\strut \small ↑12.9\%}; } \\
        & CoT & 0.537 & 0.378 & 0.058 & 0.382 & 0.119 & 0.295  \tikz[baseline=(X.base)]{\node[fill=red!50, rounded corners=2pt, inner sep=0pt] (X) {\strut \small ↓4.5\%}; } \\
        \midrule 
        \midrule

        CLIP Classifier & Baseline & 0.548 & 0.270 & \textbf{0.150} & \textbf{0.479} & \textbf{0.687} & \textbf{0.427} \\

        \bottomrule
        \end{tabularx}
        }
       \caption{Results for Coherence Relation Prediction on CLUE Single-Label. The Coherence Relations predicted are Visible, Subjective (Subj), Action, Story and Meta}
        \label{table:metrics_clue_sl}
\end{table}

\section{Experiments}
To answer our research questions, we conduct experiments on the {\name} benchmark with top open-source and proprietary MLLMs. For (RQ1), we evaluate the performance of 12 MLLMs from 9 different model families across our benchmark along with a classifier baseline. The 4 settings in our benchmark are structured with increasing difficulty, with DisRel and Tweet Subtitles being the simpler settings while CLUE Single-Label (SL) and CLUE Multi-Label (ML) are more complex. To answer (RQ2), we pick a selection of MLLMs and investigate their ability to verify coherence relations as correct or incorrect when provided along with image-text pairs. This provides a measure of the model's grasp of concepts such as discourse coherence and intermodal reasoning. For understanding (RQ3), we evaluate the effectiveness of different prompting strategies in enabling these MLLMs to discern coherence relations. We also fine-tune an MLLM on our benchmark to see if it can enhance its intermodal reasoning capability.

\begin{table}[!ht]
    \scriptsize
    \centering
        \begin{tabularx}{\linewidth}{@{} c|c|Y|Y|Y @{}}
        \toprule
        \textbf{Dataset} & \textbf{CR} & \textbf{Claude} & \textbf{Gemini} & \textbf{GPT4o} \\
        \midrule

        \multirow{3}{0.1\linewidth}{\centering DisREL} & Similar & 70.4\% & 57.2\% & 14.8\% \\
        & Complementary & 91.2\% & 10.8\% & 96.8\% \\
        & Overall & \textbf{80.8\%} & 34.0\% & 55.8\% \\
        \midrule

        \multirow{6}{0.1\linewidth}{\centering Tweet Subtitles} & Insertion & 20.59\% & 0.0\% & 11.76\% \\
        & Concretization & 74.1\% & 57.35\% & 37.61\% \\
        & Projection & 81.82\% & 0.0\% & 15.91\% \\
        & Restatement & 65.73\% & 64.34\% & 21.68\% \\
        & Extension & 66.29\% & 0.0\% & 38.29\% \\
        & Overall & \textbf{70.44\%} & 47.69\% & 34.56\% \\
        \midrule

        \multirow{6}{0.1\linewidth}{\centering CLUE SL} & Visible & 83.37\% & 90.21\% & 75.4\% \\
        & Subjective & 58.0\% & 20.0\% & 52.0\% \\
        & Action & 72.73\% & 9.09\% & 54.55\% \\
        & Story & 29.12\% & 3.85\% & 35.71\% \\
        & Meta & 9.98\% & 0.0\% & 0.8\% \\
        & Overall & \textbf{42.77\%} & 35.0\% & 36.52\% \\
        \midrule

        \multirow{2}{0.1\linewidth}{\centering CLUE ML} & \multirow{2}{*}{Overall} & \multirow{2}{*}{\textbf{ 48.82\%}} & \multirow{2}{*}{32.71\%} & \multirow{2}{*}{44.21\%} \\
        & & & & \\
        
        \bottomrule
        \end{tabularx}
       \caption{Accuracy of MLLMs in verifying each Coherence Relation (CR) of every dataset.}
        \label{table:metrics_verification}
\end{table}

\subsection{Models Evaluated} \label{models-evaluated}
We evaluate \textbf{4 proprietary MLLMs}: GPT-4o \cite{OpenAI2024-hr}, Gemini 1.5 Flash \cite{Pichai2024-xj}, Gemini 1.5 Pro \cite{Pichai2024-xj}, and Claude 3.5 Sonnet v2 \cite{AnthropicUnknown-hu} and \textbf{8 open-source MLLMs:} LLaVA 1.6 (7B, 13B, 34B) \cite{Liu2024-il}, LLaVA OneVision 7B \cite{Li2025-ad}, Qwen2-VL-7B \cite{Wang2024-fa}, Llama 3.2 11B Instruct \cite{Meta-AIUnknown-ot}, Phi3.5 Vision Instruct \cite{Abdin2024-vk}, and InternVL 2.5 26B \cite{Chen2024-yg}. We selected these model families as they demonstrated acceptable prompt adherence as described in Appendix Sections \ref{appendix-availability}, \ref{appendix-evaluation}. We also include a pre-trained classifier fine-tuned for the task of coherence relation prediction. We selected GPT-4o, Gemini 1.5 Pro, and Claude 3.5 Sonnet v2 as they were among the better-performing MLLMs on our benchmark for verification, with more details provided in Appendix Section \ref{appendix-verification}.

\subsection{Evaluation Metrics} \label{eval-metrics}
On the task of coherence relation prediction, we report the per-class F1 score and overall F1 score across all 4 settings. We select Macro F1 for overall performance as it treats all classes equally, which is important for our benchmark as it contains imbalanced classes. We report response accuracy for measuring performance on the verification task.

\subsection{Prompting Strategies and Fine-tuning} \label{prompt-strategies}
In addition to zero-shot evaluation, we also investigate the contribution of few-shot and Chain-of-Thought (CoT) prompting strategies in enabling MLLMs to learn coherence relations better. For few-shot, we include one example per coherence relation in each prompt as examples in the 3 single-label classification settings. For multi-label classification on CLUE ML, we include 6 different examples covering different combinations of relations in our prompt. To perform CoT, we include a reasoning step in our prompt that asks the model to generate a rationale before predicting the coherence relation. More details about the prompt templates used for each of the tasks are present in Sections \ref{appendix-eval-prompts} and \ref{appendix-verify-prompts} of our appendix. We fine-tune the Llama 3.2 11B Instruct model on our benchmark to measure the impact of task-specific fine-tuning in open-source MLLMs with hyperparameter selection described in Appendix Section \ref{appendix-finetuning}.

\section{Findings and Implications}
\subsection{Main Results}
\paragraph{MLLMs Struggle with Coherence Relations}
From our results in Tables \ref{table:metrics_disrel}, \ref{table:metrics_tweets}, \ref{table:metrics_clue_sl}, \ref{table:metrics_clue_ml} we observe that no MLLM shows improvements over our baseline classifier on Macro F1 scores across all settings. When strictly looking at zero-shot prompts, Claude 3.5 Sonnet v2 performs the best on Tweet Subtitles, CLUE ML, and CLUE SL while Gemini 1.5 Flash performs the best on DisRel. However, the CLIP Classifier can outperform these MLLMs by 2.4\% on DisRel, 64.1\% on Tweet Subtitles, 38.6\% on CLUE SL, and 5.6\% on CLUE ML in terms of Macro F1 score. This shows that although these datasets have clearly discernible visual and text features that help in predicting coherence relations, MLLMs aren't able to comprehend them effectively. The trend extends to both proprietary and open-source MLLMs regardless of their size. Our results reiterate the need for benchmarks such as {\name} to evaluate the intermodal reasoning capabilities of MLLMs.

\paragraph{Pragmatic Relations are Challenging}
In single-label prediction settings, we observe that MLLMs come close to the baseline classifier's scores on DisRel, containing the image-text relations that are more literal (Similar, Complementary). On the other hand, there exists a significant gap in performance in other single-label datasets. Looking into per-relation F1 scores, pragmatic relation categories such as Insertion, Projection, and Extension are particularly challenging for MLLMs. A similar trend is observed in CLUE SL and CLUE ML where MLLMs struggle with relation categories such as Story and Meta.

\paragraph{Verification Accuracy Depends on Settings}
Analyzing the verification performance of MLLMs in Table \ref{table:metrics_verification}, we observe that the performance of MLLMs on the verification task is highly dependent on the setting. Across all settings, Claude 3.5 Sonnet v2 performs the best, with an accuracy of 80.8\% on DisRel, 70.4\% on Tweet Subtitles, 42.8\% on CLUE SL and 48.5\% on CLUE ML. This shows that MLLMs are able to verify coherence relations better in settings where the relations are more literal and easier to understand. However, the performance of MLLMs on the verification task is significantly lower in settings where the relations are more non-literal and pragmatic. 

\paragraph{Inconsistency of Prompting Strategies}
In our experiments with few-shot and CoT prompting strategies, we observe that the performance of MLLMs is inconsistent across different settings and model families. Across DisRel, Tweet Subtitles, CLUE SL and CLUE ML, a total of 7, 8, 10 and 10 MLLMs respectively show improvements in performance with either few-shot or CoT prompting strategies. However, only 2 MLLMs: LLaVA OneVision 7B and GPT-4o show improvements across all settings. Overall, we observe that in the more difficult settings (CLUE SL and CLUE ML), more number of models are able to leverage one of these alternate prompting strategies to improve their performance. But, even with additional examples or reasoning steps, MLLMs are not able to outperform the baseline classifier. This shows that Coherence Relation Prediction is a fundamentally difficult task that cannot be taught to MLLMs only through prompting strategies.

\paragraph{Fine-tuning Improves MLLM Reasoning}
Looking at Figure \ref{fig:finetuning}, we observe that fine-tuning the Llama 3.2 Vision model on our benchmark proves beneficial for coherence relation prediction. In both DisRel and Tweet Subtitles, we see gains in both zero-shot and few-shot prompt scores with Llama 3.2 Vision up to 18.42\% compared to its original performance. On both CLUE ML and SL, we see improvements in either zero-shot or few-shot performance with minimal performance loss on the other. This shows that MLLMs are able to learn to recognize coherence relations better when fine-tuned on a task-specific dataset. Coherence-aware fine-tuning can be a promising direction for improving their reasoning and cognition abilities.

\subsection{Discussion}

\paragraph{Model Biases Inhibit Prediction Performance}
Looking at the per-class F1 scores across MLLMs, we observe they are biased towards certain relation categories. This includes the prediction of only a small subset of relations across all samples in an evaluation setting. From Figure \ref{fig:data_dist}, we acknowledge that the distribution of relation categories in our benchmark is imbalanced. However, this response imbalance of MLLMs is observed even on majority classes such as Concretization in Tweet Subtitles and Meta relations in CLUE SL and ML. This shows that despite providing few-shot examples and prompt optimization strategies, MLLMs display biases towards certain relation categories. When we look at the results of our fine-tuned model, we can see that prediction results on relations ignored by the base model are improved. This shows that fine-tuning can help mitigate these reasoning biases in MLLMs.

\paragraph{Cross-Discourse Generalization of CR Taxonomies}
With our evaluation of MLLMs performing MDA, we show that they perform much worse compared to baseline classifiers within each discourse type. Since CR taxonomies are designed specifically for the discourses analyzed, it is difficult to directly infer the cross-domain translation of model performances from one discourse to another. Previous studies on text-only discourses report poor cross-domain adaptation of traditional classifiers across discourse types \cite{Bourgonje2024-ia}. As {\name} extends evaluation to multiple discourses, we are able to provide a better assessment of MLLM capabilities in Discourse Analysis. A natural extension of our benchmark would be designing a unified set of CRs that can be applied across complementary discourses. This setting would be especially challenging for our classifier baselines due to the varying distribution of images and text from different sources, compared to MLLMs which are more robust to these changes. The grouping of complementary discourse domains and definition of new unified CRs is a challenging task which we aim to investigate as a part of our future work. 
\section{Conclusions}

We propose {\name}, a novel benchmark to evaluate how MLLMs perform MDA using Coherence Relations. Our experiments show existing state-of-the-art MLLMs struggle to match simple baseline classifiers in predicting Coherence Relations across different discourse domains. We also show the impact of evaluating different prompt strategies and the importance of using diverse datasets to probe intermodal reasoning capabilities of MLLMs. Finally, we show that fine-tuning MLLMs on coherence relations can help alleviate model biases and improve their performance on these tasks. This work highlights the need for MLLM benchmarks to evolve beyond factual \& perceptual assessment tasks and focus on understanding both literal and pragmatic relationships between multimodal components of real-world discourses. We hope that {\name} will serve as a stepping stone for future research in MDA and encourage the community to explore new methods to improve MLLMs on these tasks.

\section*{Limitations}
While our proposed benchmark provides a comprehensive assessment of intermodal reasoning in current MLLMs, several limitations must be acknowledged. The benchmark is currently limited to analyzing coherence relations in single-turn discourses. This is due to a lack of publicly available datasets that provide multi-turn image-text pairs with annotated coherence relations. We plan to extend our benchmark to include multi-turn discourse relations as future work. Our benchmark is currently limited to the English language and must be extended to multi-lingual discourses as well.

\section*{Acknowledgments}
This research was in part supported by the U.S. National Science Foundation (NSF) award \#1820609. Part of the research results were obtained using the computational resources provided by CloudBank (\url{https://www.cloudbank.org/}), which was supported by the NSF award \#1925001.

\bibliography{acl_latex}

\clearpage
\appendix
\section*{Appendix}
\label{sec:appendix}

\begin{table}[!ht]
    \centering
    \scalebox{0.50}{
        \begin{tabularx}{1.97\linewidth}{@{} l|Y| Y @{} Y @{} Y @{} Y @{} Y| l @{}}
        \toprule
        \textbf{Model} & \textbf{Prompt} & \textbf{Visible} & \textbf{Subj} & \textbf{Action} & \textbf{Story} & \textbf{Meta} & \textbf{Macro F1} \\
        \midrule

        \multirow{2}{*}{LLaVA 1.6 7B} & Zero & 0.864 & 0.117 & 0.113 & 0.048 & 0.029 & 0.234 \\
        & CoT & 0.848 & 0.245 & 0.247 & 0.058 & 0.013 & 0.282  \tikz[baseline=(X.base)]{\node[fill=green!50, rounded corners=2pt, inner sep=0pt] (X) {\strut \small ↑20.5\%}; } \\
        \midrule

        \multirow{2}{*}{LLaVA 1.6 13B} & Zero & 0.869 & 0.147 & 0.389 & 0.115 & 0.401 & 0.384 \\
        & CoT & 0.849 & 0.095 & 0.237 & 0.090 & 0.048 & 0.264  \tikz[baseline=(X.base)]{\node[fill=red!50, rounded corners=2pt, inner sep=0pt] (X) {\strut \small ↓31.2\%}; } \\
        \midrule

        \multirow{3}{*}{LLaVA 1.6 34B} & Zero & 0.868 & 0.165 & 0.470 & 0.369 & 0.298 & 0.434 \\
        & Few & 0.859 & 0.000 & 0.471 & 0.453 & 0.166 & 0.390  \tikz[baseline=(X.base)]{\node[fill=red!50, rounded corners=2pt, inner sep=0pt] (X) {\strut \small ↓10.1\%}; } \\
        & CoT & 0.858 & 0.117 & 0.317 & 0.175 & 0.163 & 0.326  \tikz[baseline=(X.base)]{\node[fill=red!50, rounded corners=2pt, inner sep=0pt] (X) {\strut \small ↓24.9\%}; } \\
        \midrule

        \multirow{3}{*}{LLaVA OneVision 7B} & Zero & 0.820 & 0.034 & 0.380 & 0.024 & 0.000 & 0.252 \\
        & Few & 0.757 & 0.109 & 0.510 & 0.150 & 0.000 & 0.305  \tikz[baseline=(X.base)]{\node[fill=green!50, rounded corners=2pt, inner sep=0pt] (X) {\strut \small ↑21.0\%}; } \\
        & CoT & 0.856 & 0.150 & 0.349 & 0.213 & 0.154 & 0.345  \tikz[baseline=(X.base)]{\node[fill=green!50, rounded corners=2pt, inner sep=0pt] (X) {\strut \small ↑36.9\%}; } \\
        \midrule

        \multirow{3}{*}{Qwen2-VL 7B} & Zero & 0.864 & 0.045 & 0.211 & 0.086 & 0.013 & 0.244 \\
        & Few & 0.864 & 0.162 & 0.461 & 0.368 & 0.017 & 0.374  \tikz[baseline=(X.base)]{\node[fill=green!50, rounded corners=2pt, inner sep=0pt] (X) {\strut \small ↑53.3\%}; } \\
        & CoT & 0.865 & 0.082 & 0.094 & 0.080 & 0.021 & 0.228  \tikz[baseline=(X.base)]{\node[fill=red!50, rounded corners=2pt, inner sep=0pt] (X) {\strut \small ↓6.6\%}; } \\
        \midrule

        \multirow{3}{*}{Llama 3.2 Vision 11B} & Zero & 0.869 & 0.157 & 0.424 & 0.349 & 0.284 & 0.417 \\
        & Few & 0.828 & 0.248 & 0.571 & 0.443 & \underline{0.499} & 0.518  \tikz[baseline=(X.base)]{\node[fill=green!50, rounded corners=2pt, inner sep=0pt] (X) {\strut \small ↑24.2\%}; } \\
        & CoT & 0.850 & 0.183 & 0.391 & 0.420 & 0.371 & 0.443  \tikz[baseline=(X.base)]{\node[fill=green!50, rounded corners=2pt, inner sep=0pt] (X) {\strut \small ↑6.2\%}; } \\
        \midrule

        \multirow{3}{*}{Phi3.5 Vision 4.2B} & Zero & 0.866 & 0.000 & 0.092 & 0.036 & 0.013 & 0.201 \\
        & Few & 0.527 & 0.226 & 0.311 & 0.490 & 0.036 & 0.318  \tikz[baseline=(X.base)]{\node[fill=green!50, rounded corners=2pt, inner sep=0pt] (X) {\strut \small ↑58.2\%}; } \\
        & CoT & 0.819 & 0.047 & 0.475 & 0.294 & 0.064 & 0.340  \tikz[baseline=(X.base)]{\node[fill=green!50, rounded corners=2pt, inner sep=0pt] (X) {\strut \small ↑69.2\%}; } \\
        \midrule

        \multirow{3}{*}{InternVL 2.5 26B} & Zero & 0.822 & 0.291 & 0.448 & 0.324 & 0.029 & 0.383 \\
        & Few & 0.496 & 0.266 & 0.491 & 0.400 & 0.128 & 0.356  \tikz[baseline=(X.base)]{\node[fill=red!50, rounded corners=2pt, inner sep=0pt] (X) {\strut \small ↓7.0\%}; } \\
        & CoT & 0.757 & 0.397 & 0.444 & 0.331 & 0.059 & 0.397  \tikz[baseline=(X.base)]{\node[fill=green!50, rounded corners=2pt, inner sep=0pt] (X) {\strut \small ↑3.7\%}; } \\
        \midrule 
        \midrule

        \multirow{3}{*}{GPT-4o} & Zero & 0.858 & 0.451 & 0.453 & 0.291 & 0.060 & 0.423 \\
        & Few & 0.874 & 0.495 & 0.561 & 0.525 & 0.123 & 0.515  \tikz[baseline=(X.base)]{\node[fill=green!50, rounded corners=2pt, inner sep=0pt] (X) {\strut \small ↑21.7\%}; } \\
        & CoT & 0.865 & 0.506 & 0.357 & 0.354 & 0.084 & 0.433  \tikz[baseline=(X.base)]{\node[fill=green!50, rounded corners=2pt, inner sep=0pt] (X) {\strut \small ↑2.4\%}; } \\
        \midrule

        \multirow{3}{*}{Gemini 1.5 Flash} & Zero & 0.875 & 0.368 & 0.554 & 0.355 & 0.065 & 0.443 \\
        & Few & 0.847 & 0.420 & 0.648 & 0.480 & 0.163 & 0.512  \tikz[baseline=(X.base)]{\node[fill=green!50, rounded corners=2pt, inner sep=0pt] (X) {\strut \small ↑15.6\%}; } \\
        & CoT & 0.871 & 0.419 & 0.308 & 0.358 & 0.109 & 0.413  \tikz[baseline=(X.base)]{\node[fill=red!50, rounded corners=2pt, inner sep=0pt] (X) {\strut \small ↓6.8\%}; } \\
        \midrule

        \multirow{3}{*}{Gemini 1.5 Pro} & Zero & 0.884 & 0.485 & 0.544 & 0.313 & 0.106 & 0.467 \\
        & Few & 0.866 & \underline{0.532} & \underline{0.668} & 0.464 & 0.206 & 0.547  \tikz[baseline=(X.base)]{\node[fill=green!50, rounded corners=2pt, inner sep=0pt] (X) {\strut \small ↑17.1\%}; } \\
        & CoT & 0.880 & 0.403 & 0.180 & 0.278 & 0.090 & 0.366  \tikz[baseline=(X.base)]{\node[fill=red!50, rounded corners=2pt, inner sep=0pt] (X) {\strut \small ↓21.6\%}; } \\
        \midrule

        \multirow{3}{*}{Claude 3.5 Sonnet v2} & Zero & \underline{0.891} & \textbf{0.535} & \textbf{0.681} & 0.479 & 0.220 & 0.561 \\
        & Few & 0.829 & 0.503 & 0.643 & \underline{0.553} & 0.360 & \underline{0.578}  \tikz[baseline=(X.base)]{\node[fill=green!50, rounded corners=2pt, inner sep=0pt] (X) {\strut \small ↑3.0\%}; } \\
        & CoT & 0.876 & 0.515 & 0.596 & 0.389 & 0.174 & 0.510  \tikz[baseline=(X.base)]{\node[fill=red!50, rounded corners=2pt, inner sep=0pt] (X) {\strut \small ↓9.1\%}; } \\
        \midrule 
        \midrule

        CLIP Classifier & Baseline & \textbf{0.905} & 0.176 & 0.627 & \textbf{0.615} & \textbf{0.642} & \textbf{0.593} \\

        \bottomrule
        \end{tabularx}
    }
       \caption{Results for Coherence Relation Prediction on the CLUE Multi-Label dataset. The Coherence Relations predicted are Visible, Subjective (Subj), Action, Story and Meta with multiple relations being applicable to a single image-text pair.}
        \label{table:metrics_clue_ml}
\end{table}

\begin{figure}
    \centering
    \includegraphics[width=\linewidth]{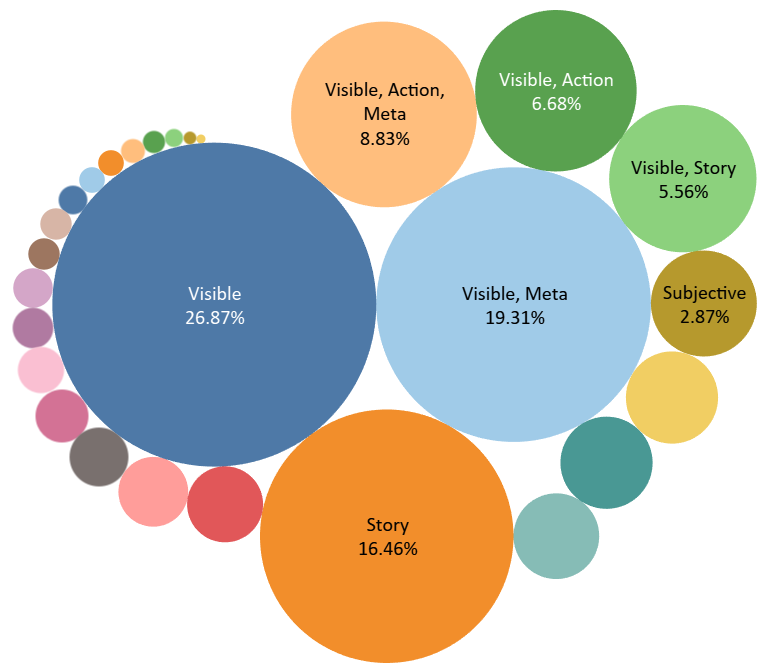}
    \caption{An overview of the Image-Text Label (i.e., Coherence Relations) distribution across CLUE ML}
    \label{fig:multilabel-distribution}
\end{figure}

\section{Data Preparation}
\label{appendix-data-prep}

This section sheds light on the methods used while preparing all the datasets mentioned in this paper for model evaluation. We verify all three datasets used to construct this benchmark have a permissive license that allows usage for research purposes without restrictions (DisRel - MIT License, Tweet Subtitles - MIT License, CLUE - Sourced from Conceptual Captions and free for research use).

\subsection{DisREL}
Due to limited number of samples in the \textbf{Unrelated} category, these image-text pairs were discarded from our train and test set. All placeholder instances of \texttt{<URL>} were removed from the text as a part of our data cleaning.

\subsection{Tweet Subtitles}
This dataset contains two types of captions for tweets: actual and text generated by an image captioning model. We use only the \textbf{actual} caption as part of our evaluation.

\subsection{CLUE}
The labels other than the ones mentioned in Section \ref{clue-labels} were disregarded from our train and test set for both settings, due to the lack of examples. We construct the CLUE Single-Label dataset with the same heuristic used by \citet{Alikhani2020-nr}:

\begin{enumerate}[leftmargin=1.25cm, label=Step \arabic*:]
    \item If the set contains a \textit{Meta} relation, assign it to the image-text pair. Else, proceed to the next step.
    \item If the set contains a \textit{Visible} relation and doesn't contain either a \textit{Meta} or \textit{Subjective} relation, assign it to the image-text pair. Else, proceed to the next step.
    \item If none of the above rules are met, randomly sample one relation from the 5 available, and assign it to the pair.
\end{enumerate}

\section{Model Availability} \label{appendix-availability}

This section focuses on the details of model availability and parameters, that we use in Section \ref{models-evaluated}. For all models, we set temperature to $0$ or \texttt{do\_sample=False}, maximum output tokens to $512$ and the random seed set to $42$, wherever possible to ensure reproducibility. The model responses in this paper were collected between January 12, 2025 and February 12, 2025.

\subsection{Proprietary Models}

\paragraph{OpenAI GPT:} We access the GPT-4o model via the official OpenAI API. We evaluate \texttt{gpt-4o-2024-08-06}.

\paragraph{Anthropic Claude:} We access Claude 3.5 Sonnet v2 via the Vertex AI API, using Google Cloud. We evaluate \texttt{claude-3-5-sonnet-v2@20241022}.

\paragraph{Google Gemini:} We access Gemini 1.5 Flash and Gemini 1.5 Pro via the Vertex AI API, using Google Cloud. We evaluate \texttt{gemini-1.5-flash-002} and \texttt{gemini-1.5-pro-002}. 

\subsection{Open Source Models}
We evaluate models published on Huggingface Hub. LLaVA 1.6 34B and Llama 3.2 11B Vision were evaluated using the LMDeploy \footnote{\url{https://github.com/InternLM/lmdeploy}} framework. We evaluate Qwen2-VL using code released by the authors. All other models, were evaluated using the VLLM \footnote{\url{https://github.com/vllm-project/vllm}} framework. Refer to Table \ref{table:mllm_ids} for the models we evaluate.

\begin{table}[!h]
    \centering
    \scriptsize
        \begin{tabularx}{\linewidth}{@{} c|*2Y|Y @{}}
        \toprule
        Model & Model ID \\
        \midrule

        InternVL 2.5 26B & \texttt{OpenGVLab/InternVL2\_5-26B} \\
        Llama 3.2 Vision 11B & \texttt{meta-llama/Llama-3.2-11B-Vision-Instruct} \\
        LLaVA 1.6 7B & \texttt{llava-hf/llava-v1.6-mistral-7b-hf} \\
        LLaVA 1.6 13B & \texttt{llava-hf/llava-v1.6-vicuna-13b-hf} \\
        LLaVA 1.6 34B & \texttt{liuhaotian/llava-v1.6-34b} \\
        LLaVA OneVision 7B & \texttt{llava-hf/llava-onevision-qwen2-7b-ov-hf} \\
        Phi 3.5 Vision & \texttt{microsoft/Phi-3.5-vision-instruct} \\
        Qwen2-VL-7B & \texttt{Qwen/Qwen2-VL-7B-Instruct} \\

        \midrule
        Claude 3.5 Sonnet v2 & \texttt{claude-3-5-sonnet-v2@20241022} \\
        GPT-4o & \texttt{gpt-4o-2024-08-06} \\
        Gemini 1.5 Flash & \texttt{gemini-1.5-flash-002} \\
        Gemini 1.5 Pro & \texttt{gemini-1.5-pro-002} \\
        
        \bottomrule
        \end{tabularx}
       \caption{MLLMs we evaluate in this paper. For open-source models, this table shows the model names in Huggingface.}
        \label{table:mllm_ids}
\end{table}

\section{MLLM Evaluation Details} \label{appendix-evaluation}
This section provides details about the \textit{evaluation} task (RQ1) mentioned in Section \ref{models-evaluated}. \\

\subsection{Prompt Templates} \label{appendix-eval-prompts}
As mentioned in Section \ref{prompt-strategies}, we make use of Zero-Shot, Few-Shot and Chain of Thought prompting for evaluation. Every prompting strategy utilizes three different messages:
\begin{itemize}
    \item \textbf{System Message:} We explain the task and the definitions of each Coherence Relation present in the dataset being evaluated.
    \item \textbf{User Message:} This message is used to reiterate the task again, along with the required output format. The image and text that needs to be evaluated, is also added here.
    \item \textbf{Assistant Message:} We use this optional message for certain models, to guide its responses towards the intended output format.
\end{itemize}

The different prompts and system messages used on each data source as mentioned in Section \ref{data-sources}, is present in the appendix.

\subsection{Few Shot Prompting}

In this prompting strategy, we utilize user-assistant message pairs that are inserted right after the user message which specifies output format. For the Tweet Subtitles and CLUE Single-Label datasets, we utilize \textbf{5-shot examples} to include all possible coherence relations. In the case of CLUE Multi-Label and DisREL, we utilize \textbf{6-shot examples} and \textbf{2-shot examples} respectively. \\

We do not evaluate LLaVA 1.6 7B and 13B using this prompting technique, as our prompt (text + multimodal tokens) does not fit into the context length (4096) of these models. 

\subsection{Chain-of-Thought Prompting}
We instruct the model to analyze the image-text pair, before assigning a Coherence Relation in this prompting strategy. We incorporate the instruction "Let's think step by step", to make the model respond with concise sentences that detail its reasoning process.

\subsection{Preprocessing Images for Claude} \label{claude-preprocess}
We noticed that some images were above the 5 MB per file size limit imposed by Anthropic for their API. As per their recommendations, we evaluate Claude on images that are resized to 1.3 megapixels, while preserving the aspect ratio.

\subsection{Postprocessing MLLM Responses}
In the case of single-label datasets, we remove instances of the phrase "Coherence Relation:" along with other punctuation and whitespace. If there exists only one occurrence of a particular coherence relation, we use that as the prediction result for the image-text pair. \\

While working with CLUE Multi-Label responses, we remove instances of the phrase "Coherence Relations:". All valid JSON in the response is parsed using regular expressions. If the output format is comma-separated values, those responses are parsed appropriately. \\

After this, if we cannot find any valid label for an image-text pair from the MLLM's response, we discard the sample from our test set. To ensure test set consistency, we discarded around \textbf{200 samples} across all datasets and calculated the final evaluation metrics as mentioned in Section \ref{eval-metrics}.

\section{MLLM Verification Details} \label{appendix-verification}
This section provides details about the \textit{verification} task (RQ2) mentioned in Section \ref{models-evaluated}.

\subsection{Prompt Templates} \label{appendix-verify-prompts}
For this task, we utilize a Chain-of-Thought prompting strategy. Each model is given the same system message as before, but along with the image-text pair, we also give the ground truth Coherence Relation. The model is then asked to respond with a True/False answer, along with its rationale for its response. 

\subsection{Preprocessing Images for Claude}
We use the same strategy as mentioned in Section \ref{claude-preprocess}, only for the images that don't come under the file size limit.

\subsection{Postprocessing MLLM Responses}
We parse boolean values from each MLLM response, and assign \textbf{False} to an image-text pair, only if there is any occurrence of the same. For CLUE ML, we provide only overall verification accuracies since it is a multi-label verification problem.

\section{Fine-tuning Details} \label{appendix-finetuning}
We fine-tune LLaMA 3.2 Vision 11B Instruct (\texttt{unsloth/Llama-3.2-11B-Vision-Instruct} in Huggingface) using the Unsloth\footnote{\url{https://unsloth.ai/blog/vision}} framework. We opted for this framework due to its memory efficiency and rapid fine-tuning capabilities. We perform Parameter Efficient Fine-Tuning (PEFT) of all layers (Vision \& Language) and modules (Attention \& MLP) present. We use the hyperparameters mentioned in Section \ref{appendix-hyperparams} on each dataset for fine-tuning. Other parameters have been initialized to their default values.

\begin{table}[t]
    \centering
    \scriptsize
        \begin{tabularx}{\columnwidth}{@{} l | Y|Y @{} Y| l @{}}
        \toprule

        \textbf{Model} & \textbf{Prompt} & \textbf{Sim} & \textbf{Compl} & \textbf{Macro F1} \\
        \midrule

        \multirow{2}{*}{FT-Llama 3.2 Vision 11B} & Zero & 0.629 & 0.620 & \textbf{0.625} \\
        & Few & \textbf{0.673} & 0.327 & 0.500  \tikz[baseline=(X.base)]{\node[fill=red!50, rounded corners=2pt, inner sep=0pt] (X) {\strut \tiny ↓20.0\%}; } \\
        \midrule

        \multirow{2}{*}{Llama 3.2 Vision 11B} & Zero & 0.388 & \textbf{0.635} & 0.512 \\
        & Few & 0.509 & 0.479 & 0.494  \tikz[baseline=(X.base)]{\node[fill=red!50, rounded corners=2pt, inner sep=0pt] (X) {\strut \tiny ↓3.5\%}; } \\
        
        \bottomrule

        \end{tabularx}
       \caption{Per-class Coherence Relation Prediction of Fine-tuned LLama 3.2 Vision 11B (FT-Llama) on the DisRel dataset. The coherence relations predicted are Similar and Complementary.}
        \label{table:metrics_disrel_finetuned}
\end{table}

\begin{table}[!ht]
    \centering
    \scalebox{0.50}{
        \begin{tabularx}{1.97\linewidth}{@{} l|Y| Y @{} Y @{} Y @{} Y @{} Y| l @{}}
        \toprule
        \textbf{Model} & \textbf{Prompt} & \textbf{Ins} & \textbf{Concr} & \textbf{Proj} & \textbf{Restmt} & \textbf{Ext} & \textbf{Macro F1} \\
        \midrule

        \multirow{2}{*}{FT-Llama 3.2 Vision 11B} & Zero & \textbf{0.440} & \textbf{0.853} & 0.045 & 0.042 & 0.148 & 0.306 \\
        & Few & 0.231 & 0.752 & \textbf{0.213} & \textbf{0.100} & \textbf{0.254} & \textbf{0.310}  \tikz[baseline=(X.base)]{\node[fill=green!50, rounded corners=2pt, inner sep=0pt] (X) {\strut \small ↑1.3\%}; } \\
        \midrule

        \multirow{2}{*}{Llama 3.2 Vision 11B} & Zero & 0.000 & 0.779 & 0.000 & 0.093 & 0.000 & 0.175 \\
        & Few & 0.035 & 0.388 & 0.000 & 0.092 & 0.113 & 0.126  \tikz[baseline=(X.base)]{\node[fill=red!50, rounded corners=2pt, inner sep=0pt] (X) {\strut \small ↓28.0\%}; } \\
        
        \bottomrule
        \end{tabularx}
        }
       \caption{Per-class Coherence Relation Prediction of Fine-tuned LLama 3.2 Vision 11B (FT-Llama) on the Tweet Subtitles dataset. The Coherence Relations predicted are Insertion (Ins), Concretization (Concr), Projection (Proj), Restatement (Restmt) and Extension (Ext).}
        \label{table:metrics_tweets_finetuned}
\end{table}

\begin{table}[!ht]
    \centering
    \scalebox{0.50}{
        \begin{tabularx}{1.97\linewidth}{@{} l|Y| Y @{} Y @{} Y @{} Y @{} Y| l @{}}
        \toprule
        \textbf{Model} & \textbf{Prompt} & \textbf{Visible} & \textbf{Subj} & \textbf{Action} & \textbf{Story} & \textbf{Meta} & \textbf{Macro F1} \\
        \midrule

        \multirow{2}{*}{FT-Llama 3.2 Vision 11B} & Zero & \textbf{0.547} & 0.074 & 0.042 & 0.045 & 0.004 & 0.142 \\
        & Few & 0.516 & \textbf{0.230} & 0.053 & \textbf{0.228} & \textbf{0.155} & \textbf{0.236}  \tikz[baseline=(X.base)]{\node[fill=green!50, rounded corners=2pt, inner sep=0pt] (X) {\strut \small ↑66.2\%}; } \\
        \midrule

        \multirow{2}{*}{Llama 3.2 Vision 11B} & Zero & 0.537 & 0.136 & \textbf{0.098} & 0.023 & 0.000 & 0.159 \\
        & Few & 0.542 & 0.000 & 0.026 & 0.000 & 0.000 & 0.114  \tikz[baseline=(X.base)]{\node[fill=red!50, rounded corners=2pt, inner sep=0pt] (X) {\strut \small ↓28.3\%}; } \\

        \bottomrule
        \end{tabularx}
        }
       \caption{Per-class Coherence Relation Prediction of Fine-tuned LLama 3.2 Vision 11B (FT-Llama) on the CLUE Single-Label dataset. The Coherence Relations predicted are Visible, Subjective (Subj), Action, Story and Meta}
        \label{table:metrics_clue_sl_finetuned}
\end{table}

\begin{table}[!ht]
    \centering
    \scalebox{0.50}{
        \begin{tabularx}{1.97\linewidth}{@{} l|Y| Y @{} Y @{} Y @{} Y @{} Y| l @{}}
        \toprule
        \textbf{Model} & \textbf{Prompt} & \textbf{Visible} & \textbf{Subj} & \textbf{Action} & \textbf{Story} & \textbf{Meta} & \textbf{Macro F1} \\
        \midrule

        \multirow{2}{*}{FT-Llama 3.2 Vision 11B} & Zero & 0.864 & 0.228 & 0.520 & 0.287 & 0.431 & 0.466 \\
        & Few & 0.864 & 0.158 & \textbf{0.586} & 0.282 & \textbf{0.549} & 0.488  \tikz[baseline=(X.base)]{\node[fill=green!50, rounded corners=2pt, inner sep=0pt] (X) {\strut \small ↑4.7\%}; } \\
        \midrule

        \multirow{2}{*}{Llama 3.2 Vision 11B} & Zero & \textbf{0.869} & 0.157 & 0.424 & 0.349 & 0.284 & 0.417 \\
        & Few & 0.828 & \textbf{0.248} & 0.571 & \textbf{0.443} & 0.499 & \textbf{0.518}  \tikz[baseline=(X.base)]{\node[fill=green!50, rounded corners=2pt, inner sep=0pt] (X) {\strut \small ↑24.2\%}; } \\

        \bottomrule
        \end{tabularx}
    }
       \caption{Per-class Coherence Relation Prediction of Fine-tuned LLama 3.2 Vision 11B (FT-Llama) on the CLUE Multi-Label dataset. The Coherence Relations predicted are Visible, Subjective (Subj), Action, Story and Meta with multiple relations being applicable to a single image-text pair.}
        \label{table:metrics_clue_ml_finetuned}
\end{table}

\subsection{Hyperparameters} \label{appendix-hyperparams}
\paragraph{Common Parameters}
\begin{itemize}
    \item LoRA Parameters: \texttt{r=16}
    \item \texttt{num\_train\_epochs = 3}
    \item \texttt{warmup\_steps = 100} since our train sets are relatively small.
    \item \texttt{per\_device\_train\_batch\_size = 32}
    \item \texttt{gradient\_accumulation\_steps = 1}
    \item \texttt{dtype = torch.bfloat16}
    \item \texttt{optim = adamw\_torch}
    \item \texttt{weight\_decay = 0.01}
    \item \texttt{lr\_scheduler\_type = cosine}
\end{itemize}

\paragraph{DisREL}
\begin{itemize}
    \item LoRA Parameters: \texttt{lora\_alpha=16}
    \item Learning Rate = $1e^{-5}$ 
\end{itemize}

\paragraph{Tweet Subtitles}
\begin{itemize}
    \item LoRA Parameters: \texttt{lora\_alpha=16}
    \item Learning Rate = $1e^{-5}$
\end{itemize}

\paragraph{CLUE Single-Label}
\begin{itemize}
    \item LoRA Parameters: \texttt{lora\_alpha=16}
    \item Learning Rate = $1e^{-5}$
\end{itemize}

\paragraph{CLUE Multi-Label}
\begin{itemize}
    \item LoRA Parameters: \texttt{lora\_alpha=8}
    \item Learning Rate = $1e^{-7}$ 
\end{itemize}

\subsection{Train Set Preparation for CLUE}
During experimentation, we noticed that models fine-tuned on CLUE Single-Label and Multi-Label, tend to skew their responses towards the majority classes (Visible, Story and Meta) in the dataset. In order to curb this behavior, we decided to randomly sample \textbf{200 examples} from the CLUE Single-Label train set for these coherence relations alone. The same image-text pairs were used for the multi-label setting as well. 

\section{Baseline Classifier Details} \label{apendix-classifier}
As mentioned in Section \ref{classifier}, we employ CLIP Text and Image Encoders (\texttt{openai/clip-vit-large-patch14} in Huggingface) in a zero-shot manner to extract multi-modal embeddings. These embeddings are then concatenated together, to form a tensor of size $1536$. This multi-modal tensor is then passed through a Multi-Layer Perceptron with two hidden layers of size $512$ and $256$, along with an output layer equal to the number of Coherence Relations in each dataset. The MLP uses RELU in between each layer for introducing non-linearity, and a Dropout of $0.2$ between the first two layers. \\

A validation split of $10\%$ was created from the train sets. The DisREL, Tweet Subtitles and CLUE Single-Label classifiers were trained using the Cross Entropy Loss, whereas the CLUE Multi-Label classifier used the Binary Cross Entropy Loss along with a Sigmoid Layer. Due to the large class imbalance in CLUE Single-Label, we use a weighted loss function in that classifier alone. Every model was trained with a batch size of $32$, using the Adam Optimizer and a learning rate of $1e^{-5}$. Table \ref{table:epochs} shows the number of epochs, for which each classifier was trained in every setting.

\begin{table}[!h]
    \centering
        \begin{tabularx}{\linewidth}{@{} c|*2Y|Y @{}}
        \toprule
        \textbf{Dataset} & \textbf{Number of Epochs} \\
        \midrule

        DisREL & 15 \\
        Tweet Subtitles & 25 \\
        CLUE Single-Label & 25 \\
        CLUE Multi-Label & 50 \\

        \bottomrule
        \end{tabularx}
       \caption{Number of epochs for which each classifier was trained.}
        \label{table:epochs}
\end{table}

\section{Computational Resources}
To evaluate and fine-tune open-source models, we use 2 NVIDIA H100 80GB HBM3 and 2 NVIDIA A100 SXM4 GPUs for around two days worth of computation.

% DisREL Prompt Templates
\begin{figure*}[t]
    \centering
    \begin{tcolorbox}[title={System Message for DisREL}, after skip=0pt, boxsep=5pt, width=\textwidth]

    You are an expert linguist and your task is to predict the Coherence Relations of a given image-text pair. A coherence relation captures the structural, logical, and purposeful relationships between an image and its text, capturing the author's intent. \\
    
    These are the possible coherence relations you can assign to an image-text pair:
    
    - Similar: The image and text provide the same information and share the same focus. There exists significant overlap in information conveyed between modalities. \\
    - Complementary: The image and text do not provide the same information or share the same focus but one modality helps understand the other better.
    
    \end{tcolorbox}
\end{figure*}

% Tweet Subtitles Prompt Template
\begin{figure*}[t]
    \centering
    \begin{tcolorbox}[title={System Message for Tweet Subtitles}, colframe = blue!30, colback = blue!10, coltitle = blue!20!black, after skip=0pt, boxsep=5pt, width=\textwidth]

    You are an expert linguist and your task is to predict the Coherence Relations of a given image-text pair. A coherence relation captures the structural, logical, and purposeful relationships between an image and its text, capturing the author's intent. \\
    
    These are the possible coherence relations you can assign to an image-text pair:
    
    - Insertion: The salient object described in the image is not explicitly mentioned in the text. \\
    - Concretization: Both the text and image contain a mention of the main visual entity. \\
    - Projection: The main entity mentioned in the text is implicitly related to the visual objects present in the image. \\
    - Restatement: The text directly describes the image contents. \\
    - Extension: The image expands upon the story or idea in the text, presenting new elements or elaborations, effectively filling in narrative gaps left by the text.
    
    \end{tcolorbox}
\end{figure*}

% CLUE SL Prompt Template
\begin{figure*}[t]
    \centering
    \begin{tcolorbox}[title={System Message for CLUE Single-Label and Multi-Label}, colframe = green!30, colback = green!10, coltitle = green!20!black, after skip=0pt, boxsep=5pt, width=\textwidth]

    You are an expert linguist and your task is to predict the Coherence Relations of a given image-text pair. A coherence relation captures the structural, logical, and purposeful relationships between an image and its text, capturing the author's intent. \\
    
    These are the possible coherence relations you can assign to an image-text pair: \\
    
    - Visible: The text presents information that is intended to recognizably characterize what is depicted in the image. \\
    - Action: The text describes an extended, dynamic process of which the moment captured in the image is a representative snapshot. \\
    - Meta: The text allows the reader to draw inferences not just about the scene depicted in the image but about the production and presentation of the image itself. \\
    - Subjective: The text provides information about the speaker's reaction to, or evaluation of, what is depicted in the image. \\
    - Story: The text provides a free-standing description of the circumstances depicted in the image, analogous to including instructional, explanatory and other background relations.
    
    \end{tcolorbox}
\end{figure*}

\begin{figure*}[t]
    \centering
    \begin{tcolorbox}[title={Zero/Few Shot Prompt for DisREL, Tweet Subtitles and CLUE Single-Label}, colframe = red!30, colback = red!10, coltitle = red!20!black, after skip=0pt, boxsep=5pt, width=\textwidth]
    
    \textbf{System} \\
    <insert-system-message> \\
    
    \textbf{User} \\
    Based on provided information, predict the most applicable Coherence Relation for the next image-text pair. Output only one relation (<insert-coherence-relations>) and do not include any other information in your response. \\

\textcolor{red}{Use the format "Coherence Relation: <insert-coherence-relation>" for your response.} \\
    (Added to finetuned LLaMA 3.2 Vision's prompt in CLUE Single-Label, to enhance output format adherence.) \\
    
    \textbf{<add-few-shot-examples>} \\
    
    \textbf{<insert-image-text-pair>} \\
    
    \textbf{Assistant} \\
    Coherence Relation:
    
    \end{tcolorbox}
\end{figure*}

\begin{figure*}[t]
    \centering
    \begin{tcolorbox}[title={CoT Prompt for DisREL, Tweet Subtitles and CLUE Single-Label}, colframe = red!30, colback = red!10, coltitle = red!20!black, after skip=0pt, boxsep=5pt, width=\textwidth]
    
    \textbf{System} \\
    <insert-system-message> \\
    
    \textbf{User} \\
    Before assigning a coherence relation, let's think step by step and analyze the image-text pair in depth. \\
        
    \textbf{<insert-image-text-pair>} \\
    
    \textbf{Assistant} \\
    Analysis: <add-analysis-from-model> \\

    \textbf{User} \\
    Based on provided information, predict the most applicable Coherence Relation for the next image-text pair. Output only one relation (<insert-coherence-relations>) and do not include any other information in your response. \\

    \textbf{Assistant} \\
    Coherence Relation:
    
    \end{tcolorbox}
\end{figure*}

% CLUE ML Prompt Template

\begin{figure*}[t]
    \centering
    \begin{tcolorbox}[title={Zero/Few Shot Prompt for CLUE Multi-Label}, colframe = orange!30, colback = orange!10, coltitle = orange!20!black, after skip=0pt, boxsep=5pt, width=\textwidth]
    
    \textbf{System} \\
    <insert-system-message> \\
    
    \textbf{User} \\
    Based on provided information, predict the correct Coherence Relations for the next image-text pair. \textcolor{red}{Output them as a JSON value to the key "labels" and do not include any other information in your response.} (Default output format for all models) \\

    \textcolor{red}{Give your predicted labels as comma separated values. Do not include any other information in your response.} \\ (Alternate output format for LLaMA 3.2, Phi 3.5, Qwen2-VL and LLaVA-OneVision) \\

    \textcolor{red}{Use the format "Coherence Relation: <insert-coherence-relation>" for your response.} \\
    (Added to LLaVA 1.6 13B prompt to enhance output format adherence.) \\
    
    \textbf{<add-few-shot-examples>} \\
    
    \textbf{<insert-image-text-pair>} \\
    
    \textbf{Assistant} \\
    Coherence Relations:
    
    \end{tcolorbox}
\end{figure*}

\begin{figure*}[t]
    \centering
    \begin{tcolorbox}[title={CoT Prompt for CLUE Multi-Label}, colframe = orange!30, colback = orange!10, coltitle = orange!20!black, after skip=0pt, boxsep=5pt, width=\textwidth]
    
    \textbf{System} \\
    <insert-system-message> \\
    
    \textbf{User} \\
    Before assigning a coherence relation, let's think step by step and analyze the image-text pair in depth. \\
        
    \textbf{<insert-image-text-pair>} \\
    
    \textbf{Assistant} \\
    Analysis: <add-analysis-from-model> \\

    \textbf{User} \\
    Now, using your analysis, predict the correct Coherence Relations for the image-text pair. \textcolor{red}{Output them as a JSON value to the key "labels" and do not include any other information in your response.} (Default output format for all models) \\

    \textcolor{red}{Give your predicted labels as comma separated values. Do not include any other information in your response.} \\
    (Alternate output format for LLaMA 3.2, Phi 3.5, Qwen2-VL and LLaVA OneVision) \\

    \textcolor{red}{Use the format "Coherence Relation: <insert-coherence-relation>" for your response.} \\
    (Added to LLaVA 1.6 13B prompt to enhance output format adherence.) \\

    \textbf{Assistant} \\
    Coherence Relations:
    
    \end{tcolorbox}
\end{figure*}

% Verification task prompt template

\begin{figure*}[t]
    \centering
    \begin{tcolorbox}[title={Verification Prompt Template}, colframe = yellow!30, colback = yellow!10, coltitle = yellow!10!black, after skip=0pt, boxsep=5pt, width=\textwidth]
    
    \textbf{System} \\
    <insert-system-message> \\
    
    \textbf{User} \\
    Based on provided information, reply True (if appropriate) or False (if not appropriate) for the following image-text pair. Give your rationale behind it. \\
        
    \textbf{<insert-image-text-pair>} \\
    \textbf{<insert-coherence-relation>} \\

    \textbf{Sample Assistant Response} \\
    <True/False>

    Rationale: <model-response>
    
    \end{tcolorbox}
\end{figure*}

\end{document}